%% file: Jamil_paper_v2.tex
\documentclass[A4,11pt]{article}
\usepackage[dvips]{graphicx}
\usepackage{amsmath}
\usepackage{amsfonts}
\usepackage{array}
\usepackage{amssymb}
\usepackage{url}
\usepackage{capt-of}            
\usepackage{placeins}           
\usepackage{supertabular}       
\usepackage{rotating}           
\usepackage{booktabs}
\usepackage{enumerate}
\usepackage{subfig}             
\usepackage{lscape}             
\usepackage{lipsum}
\usepackage{mathtools}
\usepackage{vector}

\DeclarePairedDelimiter\abs{\lvert}{\rvert}%
\title{A Literature Survey of Benchmark Functions For Global Optimization Problems}

\author
{Momin Jamil$^{*\dag}$, Xin-She Yang$^{\ddag}$ 
    $^{*}$Blekinge Institute of Technology\\
    SE-37179, Karlskrona, Sweden\\
    $^{\dag}$Harman International, Cooperate Division\\
    Becker-Goering Str. 16, D-76307 Karlsbad, Germany\\
    E-mail: momin.jamil@harman.com\\
    $^{\ddag}$Middlesex University\\
    School of Science and Technology\\
    Hendon Campus, London NW4 4BT, UK\\
    E-mail: xin-she.yang@middlesex.ac.uk 
}

\date{{\bf Citation details:} \\ 
Momin Jamil and Xin-She Yang,
A literature survey of benchmark functions for global optimization problems,
{\it Int. Journal of Mathematical Modelling and Numerical Optimisation},
Vol.~4, No.~2, pp. 150--194 (2013). \\ 
DOI: 10.1504/IJMMNO.2013.055204 
}

\begin{document}

\maketitle
Test functions are important to validate and compare the performance of  optimization algorithms.
There have been many test or benchmark functions reported in the literature; however,
there is no standard list or set of benchmark functions. Ideally, test functions should have
diverse properties so that can be truly useful to test new algorithms in an unbiased way.
For this purpose, we have reviewed and compiled a rich set of 175 benchmark functions
for unconstrained optimization problems with diverse properties in terms of modality,
separability, and valley landscape. This is by far the most complete set of functions
so far in the literature, and tt can be expected this complete set of functions can
be used for validation of new optimization in the future.

\input{part1_v2}             
\input{part2_v2}             
\input{part3_ref_v2}           

\end{document}

%% file: part1_v2.tex
\section{Introduction}

The test of reliability, efficiency and validation of optimization algorithms is frequently carried out by using a chosen set of 
common standard benchmarks or test functions from the literature. The number of test functions in
most papers varied from a few to about two dozens. Ideally, the test functions used should be diverse
and unbiased, however, there is no agreed set of test functions in the literature. Therefore, the major 
aim of this paper is to review and compile the most complete set of test functions that we can find from all the available literature
so that they can be used for future validation and comparison of optimization algorithms.

For any new optimization, it is essential to validate its performance and compare with
other existing algorithms over a good set of test functions. 
A common practice followed by many researches is to compare different algorithms
on a large test set, especially when the test involves function optimization (Gordon 1993, Whitley 1996).  However, it must be noted that
effectiveness of one algorithm against others simply cannot be measured by the problems that it solves if the the set of problems
are too specialized and without diverse properties.  Therefore, in order to evaluate
an algorithm, one must identify the kind of problems where it performs better compared to others.  This helps in characterizing the type
of problems for which an algorithm is suitable. This is only possible if the test suite is large enough to include a wide variety of problems,
such as unimodal, multimodal, regular, irregular, separable, non-separable and multi-dimensional problems.


Many test functions may be scattered in different textbooks, in individual research articles or at different web sites.
Therefore, searching for a single source of test function with a wide variety of characteristics is a cumbersome 
and tedious task. The most notable attempts to assemble global optimization test
problems can be found in \cite{ALI2005, AVERICK1991, AVERICK1992, BRANIN1972, CHUNG1998, DIXON1978, DIXON1989, FLETCHER1963, FLOUDAS1999, MORE1981, POWELL1962, POWELL1964, PRICE2005, SALOMON1996, SCHWEFEL1981, SCHWEFEL1995, SUGANTHAN2005, TANG2008, TANG2010, WHITLEY1996}.
Online collections of test problems also exist, such as the GLOBAL library at the cross-entropy toolbox \cite{CET2004}, GAMS World
\cite{GAMS2000} CUTE \cite{GOULD2001}, global optimization test problems collection by Hedar \cite{HEDAR}, collection of test functions
\cite{ANDREI2008, GEATbx, KAJ, MISHRA2006_1, MISHRA2006_2, MISHRA2006_3, MISHRA2006_4, MISHRA2006_5, MISHRA2006_6, MISHRA2006_7},
a collection of continuous global optimization test problems
COCONUT \cite{NEUMAIER2003_1} and a subset of commonly used test functions \cite{YANG2010a}.
This motivates us to carry out a thorough analysis and compile a comprehensive collection of unconstrained optimization test problems.

In general, unconstrained problems can be classified into two categories: test functions and real-world problems.
Test functions are artificial problems, and can be used to evaluate the behavior of an algorithm in sometimes diverse and difficult situations.
Artificial problems may include single global minimum, single or multiple global minima in the presence of many local minima, long narrow valleys,
null-space effects and flat surfaces.  These problems can be easily manipulated and modified to test the algorithms in diverse scenarios.  On the other hand,
real-world problems originate from different fields such as physics, chemistry, engineering, mathematics etc.  These problems are hard to manipulate and may
contain complicated algebraic or differential expressions and may require a significant amount of data to compile.  A collection of real-world unstrained
optimization problems can be found in \cite{AVERICK1991, AVERICK1992}.

In this present work, we will focus on the test function benchmarks and their diverse properties such as modality and separability. 
A function with more than one local optimum is called multimodal. These functions are used to test the ability of an algorithm to 
escape from any local minimum.
If the exploration process of an algorithm is poorly designed, then it cannot search the function landscape effectively.  This, in turn, leads to an algorithm
getting stuck at a local minimum.  Multi-modal functions with many local minima are among the most difficult class of problems for many algorithms.  Functions with flat
surfaces pose a difficulty for the algorithms, since the flatness of the function does not give the algorithm any information to direct the search process
towards the minima (Stepint, Matyas, PowerSum).  Another group of test problems is formulated by separable and non-separable functions.
According to \cite{BOYER2005}, the dimensionality of the search space is an important issue with the problem.  In some functions, the area that contains that global minima are very small, when compared to the whole search space, such as Easom, Michalewicz ($m$=10) and Powell. 
For problems such as Perm, Kowalik and Schaffer,
the global minimum is located very close to the local minima.  If the algorithm cannot keep up the direction changes in the functions with a narrow curved
valley, in case of functions like Beale, Colville, or cannot explore the search space effectively, in case of function like Pen Holder,
Testtube-Holder having multiple global minima, the algoritm will fail for these kinds of problems.  Another problem that algorithms may suffer 
is the scaling problem with many orders of magnitude differences 
between the domain and the function hyper-surface \cite{JUNIOR2004}, such as Goldstein-Price and Trid.


%% file: part2_v2.tex
\section{Characteristics of Test Functions} \label{s:collection}

The goal of any global optimization (GO) is to find the best possible solutions $\textbf{x}^{*}$ from a set
$\mathbb{X}$ according to a set of criteria $F = \{f_{1}, f_{2},\cdots f_{n}\}$.  These criteria are called
objective functions expressed in the form of mathematical functions.  An objective function is a mathematical function
$f:D \subset \Re^n \rightarrow \Re$ subject to additional constraints.  The set $D$ is referred to as the set of feasible points in
a search space.  In the case of optimizing a single criterion $f$, an optimum is either its maximum or minimum.
The global optimization problems are often defined as  minimization problems, however, these problems can be easily
converted to maximization problems by negating $f$.  A general global optimum problem can be defined as follows:
\begin{eqnarray}
   \underset{\textrm{\textbf{x}}}{\textrm{minimize}} f(\textbf{x})
\end{eqnarray}
The true optimal solution of an optimization problem may be a set of
$\textbf{x}^{*} \in D $ of all optimal points in $D$, rather than a single minimum or maximum value in some cases.  There could be
multiple, even an infinite number of optimal solutions, depending on the domain of the search space.  
The tasks of any good global optimization algorithm is to find globally optimal or at least sub-optimal solutions.
The objective functions could be characterized as continuous, discontinuous, linear, non-linear, convex, non-conxex,
unimodal, multimodal, separable\footnote{In this paper, partially separable functions are also considered as separable function}
and non-separable.

According to \cite{CHUNG1998}, it is important to ask the following two questions before start solving an optimization problem;
(i) What aspects of the function landscape make the optimization process difficult? (ii) What type of
\textit{a priori} knowledge is most effective for searching particular types of function landscape?
In order to answer these questions, benchmark functions can be classified in terms of features like
modality, basins, valleys, separability and dimensionality \cite{WINSTON1992}.

\subsection{Modality}
The number of ambiguous peaks in the function landscape corresponds to the modality of a function.
If algorithms encounters these peaks during a search process, there is a tendency that the algorithm may be trapped
in one of such peaks.  This will have a negative impact on the search process, as this can direct the search
away from the true optimal solutions.

\subsection{Basins}
A relatively steep decline surrounding a large area is called a basin.
Optimization algorithms can be easily attracted to such regions.
Once in these regions, the search process of an algorithm is severely hampered.
This is due to lack of information to direct the search process towards the minimum.
According to \cite{CHUNG1998}, a basin corresponds to the plateau for a maximization problem,
and a problem can have multiple plateaus.

\subsection{Valleys}
A valley occurs when a narrow area of little change is surrounded by regions of steep descent \cite{CHUNG1998}.
As with the basins, minimizers are initially attracted to this region.  The progress of a search process of an
algorithm may be slowed down considerably on the floor of the valley.

\subsection{Separability}
The separability is a measure of difficulty of different benchmark functions.
In general, separable functions are relatively easy to solve, when compared with their
inseperable counterpart, because each
variable of a function is independent of the other variables.  If all the parameters or variables
are independent, then a sequence of $n$ independent optimization processes can be performed.
As a result, each design variable or parameter can be optimized independently.  According to \cite{SALOMON1996}, the
general condition of separability to see if the function is easy to optimize or not is given as
\begin{eqnarray}
    \frac{\partial{f(\overline{x})}}{\partial{x_{i}}} = g(x_{i})h(\overline{x}) \label{eq:separability}
\end{eqnarray}
where $g({\overline{x_{i}}})$ means any function of $x_{i}$ only and $h({\overline{x}})$ any
function of any $\overline{x}$. If this condition is satisfied, the function is called partially
separable and easy to optimize, because solutions for each $x_{i}$ can be obtained independently
of all the other parameters.  This separability condition can be illustrated by the following two examples.

For example, function $(f_{105})$ is not separable, because it does not satisfy the condition (\ref{eq:separability})
\begin{eqnarray}
    \frac{\partial{f_{105}(x_1,x_2)}}{\partial{x_1}} &=& 400(x_1^2 - x_2)x_1 - 2x_1 -2 \nonumber \\
    \frac{\partial{f_{105}(x_1,x_2)}}{\partial{x_2}} &=& -200(x_1^2 - x_2) \nonumber
\end{eqnarray}

On the other hand, the sphere function ($f_{137}$) with two variables can indeed satisfy the above
condition (\ref{eq:separability}) as shown below.
\begin{eqnarray}
    \frac{\partial{f_{137}(x_1,x_2)}}{\partial{x_1}} = 2x_1 \nonumber ~~~~~~~~ \frac{\partial{f_{137}(x_1,x_2)}}{\partial{x_2}} = 2x_2
\end{eqnarray}
where $h(x)$ is regarded as $1$.

In \cite{BOYER2005}, the formal definition of separability is given as

\begin{eqnarray}
   \underset{\textrm{${x_{1}}$,...,${x_{p}}$}}{\textrm{arg minimize}} f({x_{1},...,x_{p}}) &=& \big(\underset{\textrm{${x_{1}}$}}{\textrm{arg minimize}}f(x_{1},...),..., \nonumber \\
   & & \underset{\textrm{${x_{p}}$}}{\textrm{arg minimize}}f(...,x_{p})\big)
\end{eqnarray}
\noindent
In other words, a function of $p$ variables is called separable, if it can written as a sum of $p$ functions of just one variable \cite{BOYER2005}.  On the other hand, a function is called non-separable, if its variables show inter-relation among themselves or are not independent.  If the objective function variables are independent of each other, then the objective functions can be decomposed into sub-objective functions.  Then, each of these sub-objectives involves only one decision variable, while treating all the others as constant and can be expressed as

\begin{eqnarray}
    f(x_{1},x_{2},\cdots,x_{p}) = \sum_{i=1}^{p}f_{i}(x_{i})
\end{eqnarray}
\noindent

\subsection{Dimensionality}
The difficulty of a problem generally increases with its dimensionality.
According to \cite{WINSTON1992, YAO1996}, as the number of parameters or dimension increases,
the search space also increases exponentially.
For highly nonlinear problems, this dimensionality may be a significant barrier for
almost all optimization algorithms.

\section{Benchmark Test Functions for Global Optimization} \label{s:collection}
Now, we present a collection of $175$ unconstrained optimization test problems which can be used to validate
the performance of optimization algorithms.  The dimensions, problem domain size and optimal solution are denoted by
$D$, $Lb \leq \textbf{x}_{i} \leq Ub$ and $f(\textbf{x}^{*}) = f(x_{1},...x_{n})$, respectively.  The symbols $\textrm{Lb}$ and $\textrm{Ub}$
represent lower, upper bound of the variables, respectively.  It is worth noting that in several cases, the optimal
solution vectors and their corresponding solutions are known only as numerical approximations.

\begin{enumerate}

\item \textbf{Ackley 1 Function} \cite{BAECK1993}(Continuous, Differentiable, Non-separable, Scalable, Multimodal)
\begin{eqnarray}
	       f_{1}(x) &=& -20 e^{-0.02 \sqrt{D^{-1}\sum_{i=1}^{D}x_{i}^{2}}} - e^{D^{-1}\sum_{i=1}^{D}\cos(2 \pi x_{i})} + 20 + e \nonumber
\end{eqnarray}
subject to  $-35 \leq x_{i} \leq 35$.  The global minima is located at origin $\textbf{x}^{*} = (0,\cdots,0)$,
$f(\textbf{x}^{*}) = 0$. \\

\item \textbf{Ackley 2 Function} \cite{ACKLEY1987} (Continuous, Differentiable, Non-Separable, Non-Scalable, Unimodal)
\begin{eqnarray}
	       f_{2}(x) &=& -200 e^ {-0.02 \sqrt{x_{1}^2 + x_{2}^2}} \nonumber
\end{eqnarray}
subject to  $-32 \leq x_{i} \leq 32$.  The global minimum is located at origin $\textbf{x}^{*} = (0,0)$,
$f(\textbf{x}^{*}) = -200$. \\

\item \textbf{Ackley 3 Function} \cite{ACKLEY1987} (Continuous, Differentiable, Non-Separable, Non-Scalable, Unimodal)
\begin{eqnarray}
	       f_{3}(x) &=& 200 e^ {-0.02 \sqrt{x_{1}^2 + x_{2}^2}} + 5 e^{\textrm{cos}(3x_{1}) + \textrm{sin}(3x_{2})}\nonumber
\end{eqnarray}
subject to  $-32 \leq x_{i} \leq 32$.  The global minimum is located at $\textbf{x}^{*} = (0,\approx -0.4)$,
$f(\textbf{x}^{*}) \approx -219.1418$. \\

\item \textbf{Ackley 4 or Modified Ackley Function} (Continuous, Differentiable, Non-Separable, Scalable, Multimodal)
        \begin{eqnarray}
	       f_{4}(\textbf{x}) &=& \sum_{i=1}^D\left(e^{-0.2}\sqrt{x_{i}^2 + x_{i+1}^2} + 3\left(\textrm{cos}(2x_{i}) + \textrm{sin}(2x_{i+1})\right)\right)\nonumber
        \end{eqnarray}
    subject to  $-35 \leq x_{i} \leq 35$.  It is highly multimodal function with two global minimum close to origin

    $\textbf{x} = f(\{-1.479252,-0.739807\}, \{1.479252,-0.739807\})$, $f(\textbf{x}^{*}) = -3.917275 $. \\

\item \textbf{Adjiman Function} \cite {ADJIMAN1998}(Continuous, Differentiable, Non-Separable, Non-Scalable, Multimodal)
\begin{eqnarray}
	       f_{5}(x) &=& \textrm{cos}(x_{1})\textrm{sin}(x_{2}) - \frac{x_{1}}{(x_{2}^2 + 1)}\nonumber
\end{eqnarray}
subject to  $ -1\leq x_{1} \leq 2$, $ -1\leq x_{2} \leq 1$.  The global minimum is located at
$\textbf{x}^{*} = (2, 0.10578)$, $f(\textbf{x}^{*}) = -2.02181$. \\

\item \textbf{Alpine 1 Function} \cite {RAHNAMAYAN2007}(Continuous, Non-Differentiable, Separable, Non-Scalable, Multimodal)
\begin{eqnarray}
	       f_{6}(\textbf{x}) &=& \sum_{i=1}^D \Big\lvert{x_i \textrm{sin}(x_i) + 0.1x_{i}}\Big\rvert \nonumber
\end{eqnarray}
    subject to  $-10 \leq x_{i} \leq 10$.  The global minimum is located at origin $\textbf{x}^{*} = (0,\cdots,0)$,
    $f(\textbf{x}^{*}) = 0$. \\

\item \textbf{Alpine 2 Function} \cite{CLERC1999} (Continuous, Differentiable, Separable, Scalable, Multimodal)
\begin{eqnarray}
	       f_{7}(\textbf{x}) &=& \prod_{i=1}^D \sqrt{x_{i}} \textrm{sin}(x_{i}) \nonumber
\end{eqnarray}
    subject to  $0 \leq x_{i} \leq 10$.  The global minimum is located at  $\textbf{x}^{*} = (7.917 \cdots 7.917)$,
    $f(\textbf{x}^{*}) = 2.808^{D}$. \\

\item \textbf{Brad Function} \cite{BRAD1970} (Continuous, Differentiable, Non-Separable, Non-Scalable, Multimodal)
\begin{eqnarray}
	       f_{8}(\textbf{x}) &=& \sum_{i=1}^{15} \left[\frac{y_{i} - x_{1} - u_{i}}{v_{i}x_{2} + w_{i}x_{3}}\right]^2 \nonumber
\end{eqnarray}
where $u_{i} = i$, $v_{i} = 16 - i$, $w_{i}$ = min($u_{i},v_{i}$) and  $\textbf{\underline{y}} = y_{i}$ = $[0.14$, $0.18$, $0.22$,
$0.25, 0.29$, $0.32, 0.35$, $0.39, 0.37$, $0.58, 0.73, 0.96$, $1.34, 2.10, 4.39]^T$.
It is subject to  $-0.25 \leq x_{1} \leq 0.25$, $0.01 \leq x_{2},x_{3} \leq 2.5$.  The global minimum is located at
$\textbf{x}^{*} = (0.0824, 1.133,2.3437)$, $f(\textbf{x}^{*}) = 0.00821487$. \\

\item \textbf{Bartels Conn Function} (Continuous, Non-differentiable, Non-Separable, Non-Scalable, Multimodal)
\begin{eqnarray}
	       f_{9}(\textbf{x}) &=& \bigl \lvert {x_{1}^2 + x_{2}^2 + x_{1}x_{2}}\bigr \rvert  + \bigl \lvert {\textrm{sin}(x_{1})}\bigr \rvert +
                                \bigl \lvert {\textrm{cos}(x_{2})}\bigr \rvert\nonumber
\end{eqnarray}
subject to  $-500 \leq x_{i} \leq 500$.  The global minimum is located at  $\textbf{x}^{*} = (0,0)$,
    $f(\textbf{x}^{*}) = 1$. \\

\item \textbf{Beale Function} (Continuous, Differentiable, Non-Separable, Non-Scalable, Unimodal)
\begin{eqnarray}
    f_{10}(\textbf{x}) &=& (1.5 - x_{1} + x_{1}x_{2})^{2} + (2.25 - x_{1} + x_{1}x_{2}^2)^{2}  \nonumber \\
                                           & & + (2.625 - x_{1} + x_{1}x_{2}^{3})^{2} \nonumber
\end{eqnarray}
subject to  $-4.5 \leq x_{i} \leq 4.5$.  The global minimum is located at $\textbf{x}^{*}=(3,0.5)$, $f(\textbf{x}^{*}) = 0 $. \\



\item \textbf{Biggs EXP2 Function} \cite{BIGGS1971} (Continuous, Differentiable, Non-Separable, Non-Scalable, Multimodal)
\begin{eqnarray}
    f_{11}(\textbf{x}) &=& \sum_{i=1}^{10} \left(e^{-t_{i}x_{1}} - 5e^{-t_{i}x_{2}} - y_{i} \right)^2 \nonumber
\end{eqnarray}
where $t_{i} = 0.1i$, $y_{i}=e^{-t_{i}} - 5e^{10t_{i}}$.  It is subject to  $0 \leq x_{i} \leq 20$.
The global minimum is located at $\textbf{x}^{*}=(1,10)$, $f(\textbf{x}^{*}) = 0 $. \\

\item \textbf{Biggs EXP3 Function} \cite{BIGGS1971} (Continuous, Differentiable, Non-Separable, Non-Scalable, Multimodal)
\begin{eqnarray}
    f_{12}(\textbf{x}) &=& \sum_{i=1}^{10} \left(e^{-t_{i}x_{1}} - x_{3}e^{-t_{i}x_{2}} - y_{i} \right)^2 \nonumber
\end{eqnarray}
where $t_{i} = 0.1i$, $y_{i}=e^{-t_{i}} - 5e^{10t_{i}}$.  It is subject to  $0 \leq x_{i} \leq 20$.
The global minimum is located at $\textbf{x}^{*}=(1,10,5)$, $f(\textbf{x}^{*}) = 0 $. \\

\item \textbf{Biggs EXP4 Function} \cite{BIGGS1971} (Continuous, Differentiable, Non-Separable, Non-Scalable, Multimodal)
\begin{eqnarray}
    f_{13}(\textbf{x}) &=& \sum_{i=1}^{10} \left(x_{3}e^{-t_{i}x_{1}} - x_{4}e^{-t_{i}x_{2}} - y_{i} \right)^2 \nonumber
\end{eqnarray}
where $t_{i} = 0.1i$, $y_{i}=e^{-t_{i}} - 5e^{10t_{i}}$.  It is subject to  $0 \leq x_{i} \leq 20$.
The global minimum is located at $\textbf{x}^{*}=(1,10,1,5)$, $f(\textbf{x}^{*}) = 0 $. \\

\item \textbf{Biggs EXP5 Function} \cite{BIGGS1971} (Continuous, Differentiable, Non-Separable, Non-Scalable, Multimodal)
\begin{eqnarray}
    f_{14}(\textbf{x}) &=& \sum_{i=1}^{11} \left(x_{3}e^{-t_{i}x_{1}} - x_{4}e^{-t_{i}x_{2}} + 3e^{-t_{i}x_{5}} - y_{i} \right)^2 \nonumber
\end{eqnarray}
where $t_{i} = 0.1i$, $y_{i}=e^{-t_{i}} - 5e^{10t_{i}} + 3e^{-4t_{i}}$.  It is subject to  $0 \leq x_{i} \leq 20$.
The global minimum is located at $\textbf{x}^{*}=(1,10,1,5,4)$, $f(\textbf{x}^{*}) = 0 $. \\

\item \textbf{Biggs EXP5 Function} \cite{BIGGS1971} (Continuous, Differentiable, Non-Separable, Non-Scalable, Multimodal)
\begin{eqnarray}
    f_{15}(\textbf{x}) &=& \sum_{i=1}^{13} \left(x_{3}e^{-t_{i}x_{1}} - x_{4}e^{-t_{i}x_{2}} + x_{6}e^{-t_{i}x_{5}} - y_{i} \right)^2 \nonumber
\end{eqnarray}
where $t_{i} = 0.1i$, $y_{i}=e^{-t_{i}} - 5e^{10t_{i}} + 3e^{-4t_{i}}$.  It is subject to  $-20 \leq x_{i} \leq 20$.
The global minimum is located at $\textbf{x}^{*}=(1,10,1,5,4,3)$, $f(\textbf{x}^{*}) = 0 $. \\

\item \textbf{Bird Function} \cite{MISHRA2006_6} (Continuous, Differentiable, Non-Separable, Non-Scalable, Multimodal)
\begin{eqnarray}
    f_{16}(\textbf{x}) &=&  \textrm{sin}(x_{1})e^{(1 - \textrm{cos}(x_{2}))^2} + \textrm{cos}(x_{2})e^{(1 - \textrm{sin}(x_{1}))^2}+ (x_{1} - x_{2})^2 \nonumber
\end{eqnarray}
subject to  $-2\pi \leq x_{i} \leq 2\pi$.  The global minimum is located at $\textbf{x}^{*}=(4.70104$, $3.15294)$,$(-1.58214$, $-3.13024)$,
$f(\textbf{x}^{*}) = -106.764537 $. \\

\item \textbf{Bohachevsky 1 Function} \cite{BOHACHEVSKY1986} (Continuous, Differentiable, Separable, Non-Scalable, Multimodal)
\begin{eqnarray}
	   f_{17}(\textbf{x}) &=& x_{1}^{2} + 2x_{2}^{2} - 0.3\textrm{cos}(3 \pi x_{1}) \nonumber \\
                                           & & - 0.4\textrm{cos}(4 \pi x_{2}) + 0.7  \nonumber
\end{eqnarray}
    subject to  $-100 \leq x_{i} \leq 100$.  The global minimum is located at $\textbf{x}^{*} = f(0,0)$,
    $f(\textbf{x}^{*}) = 0$.\\

\item \textbf{Bohachevsky 2 Function} \cite{BOHACHEVSKY1986} (Continuous, Differentiable, Non-separable, Non-Scalable, Multimodal)
\begin{eqnarray}
	    f_{18}(\textbf{x}) &=& x_{1}^{2} + 2x_{2}^{2} - 0.3\textrm{cos}(3 \pi x_{1})\cdot0.4\textrm{cos}(4 \pi x_{2}) \nonumber \\
                                           & &  + 0.3  \nonumber
\end{eqnarray}
    subject to  $-100 \leq x_{i} \leq 100$.  The global minimum is located at $\textbf{x}^{*} = f(0,0)$,
    $f(\textbf{x}^{*}) = 0$.\\

\item \textbf{Bohachevsky 3 Function} \cite{BOHACHEVSKY1986} (Continuous, Differentiable, Non-Separable, Non-Scalable, Multimodal)
\begin{eqnarray}
	       f_{19}(\textbf{x}) &=& x_{1}^{2} + 2x_{2}^{2} -0.3\textrm{cos}(3 \pi x_{1} + 4 \pi x_{2}) + 0.3 \nonumber
\end{eqnarray}
    subject to  $-100 \leq x_{i} \leq 100$.  The global minimum is located at $\textbf{x}^{*} = f(0,0)$,
    $f(\textbf{x}^{*}) = 0$.\\

\item \textbf{Booth Function} (Continuous, Differentiable, Non-separable, Non-Scalable, Unimodal) 
\begin{eqnarray}
	   f_{20}(\textbf{x}) &=& (x_{1} + 2x_{2} - 7)^{2} + (2x_{1} + x_{2} - 5)^{2}\nonumber
\end{eqnarray}
subject to  $-10 \leq x_{i} \leq 10$.  The global minimum is located at $\textbf{x}^{*} = f(1,3)$,
$f(\textbf{x}^{*}) = 0$.\\

\item \textbf{Box-Betts Quadratic Sum Function} \cite{ALI2005} (Continuous, Differentiable, Non-Separable, Non-Scalable, Multimodal)
\begin{eqnarray}
	       f_{21}(\textbf{x}) &=& \sum_{i=0}^{D-1}g(x_{i})^2 \nonumber \\
           \textrm{where} \nonumber \\
                                      g(x) &=& e^{-0.1(i + 1)x_{1}} - e^{-0.1(i + 1)x_{2}} -  e^{[(-0.1(i + 1)) - e^{-(i + 1)}]x_{3}} \nonumber
\end{eqnarray}
subject to  $0.9 \leq x_{1} \leq 1.2$, $9 \leq x_{2} \leq 11.2$, $0.9 \leq x_{2} \leq 1.2$.  The global minimum is located at $\textbf{x}^{*} = f(1,10,1)$
$f(\textbf{x}^{*}) = 0$.  \\

\item \textbf{Branin RCOS Function} \cite{BRANIN1972} (Continuous, Differentiable, Non-Separable, Non-Scalable, Multimodal)
\begin{eqnarray}
    f_{22}(\textbf{x}) &=& \left(x_2 - \frac{5.1 x_1^2}{4 \pi ^2} + \frac{5 x_1}{\pi}- 6 \right)^2 \nonumber \\
                                    & & + 10 \left(1 - \frac{1}{8 \pi} \right) \cos(x_1)  + 10 \nonumber
\end{eqnarray}
with domain $-5 \leq x_1 \le 10$, $0 \leq x_1 \leq 15$. It has three global minima at $\textbf{x}^{*} = f(\{-\pi, 12.275\}, \{\pi, 2.275\}, \{3\pi, 2.425\})$,
$f(\textbf{x}^{*}) = 0.3978873$.  \nonumber \\

\item \textbf{Branin RCOS 2 Function} \cite{MUNTEANU1998} (Continuous, Differentiable, Non-Separable, Non-Scalable, Multimodal)
\begin{eqnarray}
    f_{23}(\textbf{x}) &=& \left(x_2 - \frac{5.1 x_1^2}{4 \pi ^2} + \frac{5 x_1}{\pi}- 6 \right)^2 \nonumber \\
                       & & + 10 \left(1 - \frac{1}{8 \pi} \right) \cos(x_1)\cos(x_2)  \ln(x_{1}^{2} + x_{2}^{2} + 1)+ 10 \nonumber
\end{eqnarray}
with domain $-5 \leq x_i \le 15$.  The global minimum is located at $\textbf{x}^{*} = f(-3.2, 12.53)$,
$f(\textbf{x}^{*}) = 5.559037$.  \nonumber \\

\item \textbf{Brent Function} \cite{BRANIN1972} (Continuous, Differentiable, Non-Separable, Non-Scalable, Unimodal)
\begin{eqnarray}
    f_{24}(\textbf{x}) &=& \left(x_1 + 10\right)^2 + \left(x_2 + 10\right)^2 + e^{-x_{1}^2 - x_{2}^{2}}
\end{eqnarray}
with domain $-10 \leq x_i \le 10$.  The global minimum is located at $\textbf{x}^{*} = f(0,0)$,
$f(\textbf{x}^{*}) = 0$.  \nonumber \\


\item \textbf{Brown Function} \cite{BEGAMBRE2009} (Continuous, Differentiable, Non-Separable, Scalable, Unimodal)
\begin{eqnarray}
	       f_{25}(\textbf{x}) &=& \sum_{i=1}^{n-1}(x_{i}^{2})^{(x_{i+1}^{2} + 1)} + (x_{i+1}^{2})^{(x_{i}^{2} + 1)} \nonumber
\end{eqnarray}
subject to  $-1 \leq x_{i} \leq 4$.  The global minimum is located at $\textbf{x}^{*} = f(0,\cdots,0)$, $f(\textbf{x}^{*}) = 0$.  \\

Bukin functions \cite{SILAGADZE2007} are almost fractal (with fine seesaw edges) in the surroundings of their minimal points.
Due to this property, they are extremely difficult to optimize  by  any  global or local optimization methods.

\item \textbf{Bukin 2 Function} (Continuous, Differentiable, Non-Separable, Non-Scalable, Multimodal)
\begin{eqnarray}
	       f_{26}(\textbf{x}) &=& 100(x_{2} - 0.01x_{1}^2 + 1) + 0.01(x_{1} + 10)^2  \nonumber
\end{eqnarray}
subject to  $-15 \leq x_{1} \leq -5$ and $-3 \leq x_{2} \leq -3$.  The global minimum is located at $\textbf{x}^{*} = f(-10,0)$,
$f(\textbf{x}^{*}) = 0$.  \\

\item \textbf{Bukin 4 Function} (Continuous, Non-Differentiable, Separable, Non-scalable, Multimodal)
\begin{eqnarray}
	       f_{27}(\textbf{x}) &=& 100x_{2}^2 + 0.01\|x_{1} + 10\|  \nonumber
\end{eqnarray}
subject to  $-15 \leq x_{1} \leq -5$ and $-3 \leq x_{2} \leq -3$.  The global minimum is located at $\textbf{x}^{*} = f(-10,0)$,
$f(\textbf{x}^{*}) = 0$.  \\


\item \textbf{Bukin 6 Function} (Continuous, Non-Differentiable, Non-Separable, Non-Scalable, Multimodal)
\begin{eqnarray}
	       f_{28}(\textbf{x}) &=& 100\sqrt{\|x_{2} - 0.01x_{1}^2\|} + 0.01\|x_{1} + 10 \|  \nonumber
\end{eqnarray}
subject to  $-15 \leq x_{1} \leq -5$ and $-3 \leq x_{2} \leq -3$.  The global minimum is located at $\textbf{x}^{*} = f(-10,1)$,
$f(\textbf{x}^{*}) = 0$.\\

\item \textbf{Camel Function -- Three Hump} \cite{BRANIN1972} (Continuous, Differentiable, Non-Separable, Non-Scalable, Multimodal)
\begin{eqnarray}
	       f_{29}(\textbf{x}) &=&  2x_{1}^{2} - 1.05x_{1}^{4} + x_{1}^6/6 + x_{1}x_{2} + x_{2}^{2} \nonumber
\end{eqnarray}
    subject to  $-5 \leq x_{i} \leq 5$.  The global minima is located at $\textbf{x}^{*} = f(0,0)$, $f(\textbf{x}^{*}) = 0$.

\item \textbf{Camel Function -- Six Hump} \cite{BRANIN1972} (Continuous, Differentiable, Non-Separable, Non-Scalable, Multimodal)
\begin{eqnarray}
	       f_{30}(\textbf{x}) &=&  (4 - 2.1x_{1}^{2} + \frac{x_{1}^{4}}{3})x_{1}^{2} \nonumber \\
                                           & &  + x_{1}x_{2} + (4x_{2}^{2} - 4)x_{2}^{2} \nonumber
\end{eqnarray}
    subject to  $-5 \leq x_{i} \leq 5$.  The two global minima are located at $\textbf{x}^{*}$ = $f(\{-0.0898, 0.7126\}$, $\{0.0898, -0.7126, 0\})$,
     $f(\textbf{x}^{*}) = -1.0316$.

\item \textbf{Chen Bird Function} \cite{CHEN2003} (Continuous, Differentiable, Non-Separable, Non-Scalable, Multimodal)
\begin{eqnarray}
    f_{31}(\textbf{x}) &=& -\frac{0.001}{\bigl \lfloor{(0.001)^2 + (x_1 - 0.4x_2 - 0.1)^2}\bigr \rfloor} - \nonumber \\
                       & & \frac{0.001}{\bigl \lfloor{(0.001)^2 + (2x_1 + x_2 - 1.5)^2} \bigr \rfloor}  \nonumber
\end{eqnarray}
subject to  $-500 \leq x_{i} \leq 500$ The global minimum is located at $\textbf{x}^{*} = f(-\frac{7}{18}, -\frac{13}{18})$,
$f(\textbf{x}^{*}) = -2000$.\\

\item \textbf{Chen V Function} \cite{CHEN2003} (Continuous, Differentiable, Non-Separable, Non-Scalable, Multimodal)
\begin{eqnarray}
    f_{32}(\textbf{x}) &=& -\frac{0.001}{\bigl \lfloor{(0.001)^2 + (x_1^2 + x_2^2 - 1)^2}\bigr \rfloor} - \nonumber \\
                       & &  \frac{0.001}{\bigl \lfloor{(0.001)^2 + (x_1^2 + x_2^2 - 0.5)^2}\bigr \rfloor} - \nonumber \\
                       & &  \frac{0.001}{\bigl \lfloor{(0.001)^2 + (x_1^2 - x_2^2)^2}\bigr \rfloor}         \nonumber
\end{eqnarray}
subject to  $-500 \leq x_{i} \leq 500$ The global minimum is located at $\textbf{x}^{*}$ = $f(-\-0.3888889$, $0.7222222)$,
$f(\textbf{x}^{*}) = -2000$.\\

\item \textbf{Chichinadze Function} (Continuous, Differentiable, Separable, Non-Scalable, Multimodal)
\begin{eqnarray}
    f_{33}(\textbf{x})  &=& x_{1}^{2} - 12x_{1} + 11 + \nonumber \\
                        & & 10\textrm{cos}(\pi x_{1}/2) + 8\textrm{sin}(5\pi x_{1}/2) - \nonumber \\
                        & & (1/5)^{0.5}\exp(-0.5(x_{2} - 0.5){^2}) \nonumber
\end{eqnarray}
subject to  $-30 \leq x_{i} \leq 30$.  The global minimum is located at $\textbf{x}^{*} = f(5.90133, 0.5) $,
    $f(\textbf{x}^{*}) = -43.3159$.  \\

\item \textbf{Chung Reynolds Function} \cite{CHUNG1998} (Continuous, Differentiable, Partially-Separable, Scalable, Unimodal)
\begin{eqnarray}
    f_{34}(\textbf{x})  &=& \bigl(\sum_{i=1}^{D}x_{i}^2\bigr)^2 \nonumber
\end{eqnarray}
subject to  $-100 \leq x_{i} \leq 100$.  The global minimum is located at $\textbf{x}^{*} = f(0,\cdots,0) $,
    $f(\textbf{x}^{*}) = 0$.  \\

\item \textbf{Cola Function} \cite{ADORIO2005} (Continuous, Differentiable, Non-Separable, Non-Scalable, Multimodal)

The 17-dimensional function computes indirectly the
formula $(D,u)$  by setting $x_0 = y_0, x_1 = u_0, x_i = u_{2(i-2)},
y_i = u_{2(i-2)+1}$
$$
    f_{35}(n, u) = h(x,y) = \sum_{j<i} (r_{i,j} - d_{i,j}) ^2
$$
where $r_{i,j}$ is given by
$$
    r_{i,j} = [ (x_i-x_j)^2 + (y_i - y_j)^2 ]^{1/2}
$$
and $d$ is a symmetric matrix given by
$$
\textbf{d} = [d_{ij}]= \left(
\begin{array}{llllllllll}
  1.27 \\
  1.69 &1.43 \\
  2.04 &2.35 &2.43 \\
  3.09 &3.18 &3.26 &2.85 \\
  3.20 &3.22 &3.27 &2.88 &1.55 \\
  2.86 &2.56 &2.58 &2.59 &3.12 &3.06 \\
  3.17 &3.18 &3.18 &3.12 &1.31 &1.64 &3.00 \\
  3.21 &3.18 &3.18 &3.17 &1.70 &1.36 &2.95 &1.32\\
  2.38 &2.31 &2.42 &1.94 &2.85 &2.81 &2.56 &2.91 &2.97 \\
\end{array}
\right)
$$
This function has bounds $0 \le x_0 \le 4$ and $-4 \le x_i \le 4$ for
$i = 1 \ldots D-1$. It has a global minimum of $f(\textbf{x}^{*}) = 11.7464$. \\

\item \textbf{Colville Function} (Continuous, Differentiable, Non-Separable, Non-Scalable, Multimodal)   
\begin{eqnarray}
	       f_{36}(\textbf{x}) &=& 100(x_{1} - x_{2}^{2})^{2} + (1 - x_{1})^{2} + \nonumber \\
                                           & & 90(x_{4} - x_{3}^{2})^{2} + (1 - x_{3})^{2}  + \nonumber \\
                                           & & 10.1((x_{2} - 1)^{2} + (x_{4} - 1)^{2}) + \nonumber \\
                                           & & 19.8(x_{2} - 1)(x_{4} - 1) \nonumber
\end{eqnarray}
subject to  $-10 \leq x_{i} \leq 10$.  The global minima is located at $\textbf{x}^{*} = f(1,\cdots,1)$, $f(\textbf{x}^{*}) = 0$.  \\


\item \textbf{Corana Function} \cite{CORANA1987} (Discontinuous, Non-Differentiable, Separable, Scalable, Multimodal)
\begin{eqnarray}
        f_{37}(\textbf{x}) =
        {\left\{
            \begin{array}{l}
                0.15\left( {{z_i} - 0.05{\mathop{\rm sgn}} {{\left( {{z_i}} \right)}^2}} \right){d_i}{\rm{~ ~if }}\left| {{v_i}} \right| < A \nonumber \\
                {d_i}x_i^2{\rm{~ ~ ~ ~ ~ ~ ~ ~ ~ ~ ~ ~ ~ ~ ~ ~ ~ ~ ~ ~ ~ ~ ~ ~ ~ ~ ~ ~ ~ otherwise}} \nonumber
            \end{array} \right.}
\end{eqnarray}
where
\begin{eqnarray}
    v_{i} &=& \left|x_{i} - z_{i} \right|, ~ ~ A = 0.05 \nonumber \\
    z_{i} &=& 0.2\left\lfloor {\left| {\frac{{{x_{i}}}}{{0.2}}} \right| + 0.49999} \right\rfloor {{\rm sgn}} \left( {{x_{i}}} \right) \nonumber \\
    d_{i} &=& (1,1000,10,100)
\end{eqnarray}
subject to  $-500 \leq x_{i} \leq 500$.  The global minimum is located at $\textbf{x}^{*} = f(0,0,0,0)$,
$f(\textbf{x}^{*}) = 0$.  \\

\item \textbf{Cosine Mixture Function} \cite{ALI2005} (Discontinuous, Non-Differentiable, Separable, Scalable, Multimodal)
        \begin{eqnarray}
	       f_{38}(\textbf{x}) &=& -0.1\sum_{i=1}^{n}\textrm{cos}(5 \pi x_{i}) - \sum_{i=1}^{n}x_{i}^{2} \nonumber
        \end{eqnarray}
    subject to  $-1 \leq x_{i} \leq 1$.  The global minimum is located at $\textbf{x}^{*} = f(0,0)$,
    $f(\textbf{x}^{*}) = (0.2 \, \textrm{or} \, 0.4) \, \textrm{for} \, \, \textrm{n} = 2 \, \textrm{and} \, 4$
    respectively.  \\

\item \textbf{Cross-in-Tray Function} \cite{MISHRA2006_6} (Continuous, Non-Separable, Non-Scalable, Multimodal)
        \begin{eqnarray}
	       f_{39}(\textbf{x}) &=& -0.0001[|\textrm{sin}(x_{1})\textrm{sin}(x_{2})  \nonumber \\
                                           & & e^{|100 - [(x_{1}^{2} + x_{2}^{2})]^{0.5}/\pi|}| + 1]^{0.1} \nonumber
        \end{eqnarray}
    subject to  $-10 \leq x_{i} \leq 10$.

    The four global minima are located at $\textbf{x}^{*}$ =
    $f(\pm 1.349406685353340$, $\pm 1.349406608602084)$,
    $f(\textbf{x}^{*}) = -2.06261218 $. \\

\item \textbf{Csendes Function} \cite{CSENDES1997} (Continuous, Differentiable, Separable, Scalable, Multimodal)
\begin{eqnarray}
	       f_{40}(\textbf{x}) &=& \sum_{i=1}^D x_{i}^6\left(2 + \sin\frac{1}{x_{i}}\right) \nonumber
\end{eqnarray}
    subject to  $-1 \leq x_{i} \leq 1$.  The global minimum is located at $\textbf{x}^{*} = f(0,\cdots,0)$,
    $f(\textbf{x}^{*}) = 0$.  \\

\item \textbf{Cube Function} \cite{LAVI1966} (Continuous, Differentiable, Non-Separable, Non-Scalable, Unimodal)
\begin{eqnarray}
    f_{41}(\textbf{x}) &=& 100\left(x_2 - x_{1}^3\right)^2 + (1 - x_{1})^2 \nonumber
\end{eqnarray}
subject to  $-10 \leq x_{i} \leq 10$.  The global minimum is located at $\textbf{x}^{*} = f(-1,1)$,
    $f(\textbf{x}^{*}) = 0 $. \\

\item \textbf{Damavandi Function} \cite{DAMAVANDI2005} (Continuous, Differentiable, Non-Separable, Non-Scalable, Multimodal)
        \begin{eqnarray}
	       f_{42}(\textbf{x}) &=& \left[1 - {\left|{\frac{\textrm{sin}[\pi(x_{1} - 2)]\textrm{sin}[\pi(x_{2} - 2)]}{\pi^2(x_{1} - 2)(x_{2}-2)}}\right|}^5\right]  \nonumber \\
                                           & & \left[2 + (x_{1} - 7)^{2} + 2(x_{2} - 7 )^{2}\right] \nonumber
        \end{eqnarray}
    subject to  $0 \leq x_{i} \leq 14$.  The global minimum is located at $\textbf{x}^{*} = f(2,2)$,
    $f(\textbf{x}^{*}) = 0$. \\

\item \textbf{Deb 1 Function} (Continuous, Differentiable, Separable, Scalable, Multimodal)
\begin{eqnarray}
	       f_{43}(\textbf{x}) &=& -\frac{1}{D}\sum_{i=1}^{D}\textrm{sin}^{6}(5\pi x_{i}) \nonumber
        \end{eqnarray}
    subject to  $-1 \leq x_{i} \leq 1$.  The number of global minima is $5^{D}$ that are evenly spaced in the function landscape,
    where $D$ represents the dimension of the problem. \\

\item \textbf{Deb 3 Function} (Continuous, Differentiable, Separable, Scalable, Multimodal)
\begin{eqnarray}
	       f_{44}(\textbf{x}) &=& -\frac{1}{D}\sum_{i=1}^{D}\textrm{sin}^{6}(5\pi(x_{i}^{3/4} - 0.05)) \nonumber
        \end{eqnarray}
    subject to  $-1 \leq x_{i} \leq 1$.  The number of global minima is $5^{D}$ that are unevenly spaced in the function landscape,
    where $D$ represents the dimension of the problem. \\

\item \textbf{Deckkers-Aarts Function} \cite{ALI2005} (Continuous, Differentiable, Non-Separable, Non-Scalable, Multimodal)
\begin{eqnarray}
	       f_{45}(\textbf{x}) &=& 10^{5}x_{1}^{2} + x_{2}^{2} - (x_{1}^{2} + x_{2}^{2})^{2} +
                                            10^{-5}(x_{1}^{2} + x_{2}^{2})^{4} \nonumber
        \end{eqnarray}
    subject to  $-20 \leq x_{i} \leq 20$.  The two global minima are located at $\textbf{x}^{*} = f(0, \pm 15)$
    $f(\textbf{x}^{*}) = -24777 $. \\

\item \textbf{deVilliers Glasser 1 Function} \cite{deVILLERS1981}(Continuous, Differentiable, Non-Separable, Non-Scalable, Multimodal)
\begin{eqnarray}
    f_{46}(\textbf{x}) &=& \sum_{i=1}^{24}\left[x_{1}x_{2}^{t_{i}} \sin(x_{3}t_{i} + x_{4}) - y_{i}\right]^2 \nonumber
\end{eqnarray}
where $t_{i} = 0.1(i - 1)$, $y_{i} = 60.137\times1.371^{t_{i}}\sin(3.112t_{i} + 1.761)$.  It is
subject to  $-500 \leq x_{i} \leq 500$.  The global minimum is $f(\textbf{x}^{*}) = 0 $. \\

\item \textbf{deVilliers Glasser 2 Function} \cite{deVILLERS1981} (Continuous, Differentiable, Non-Separable, Non-Scalable, Multimodal)
\begin{eqnarray}
    f_{47}(\textbf{x}) &=& \sum_{i=1}^{16}\left[x_{1}x_{2}^{t_{i}} \tanh\left[x_{3}t_{i} + \sin(x_{4}t_{i})\right]\cos(t_{i}e^{x_{5}}) - y_{i}\right]^2 \nonumber
\end{eqnarray}
where $t_{i} = 0.1(i - 1)$, $y_{i} = 53.81\times1.27^{t_{i}}\tanh(3.012t_{i} + \sin(2.13t_{i}))\cos(e^{0.507}t_{i})$.  It is
subject to  $-500 \leq x_{i} \leq 500$.  The global minimum is $f(\textbf{x}^{*}) = 0 $. \\

\item \textbf{Dixon \& Price Function} \cite{DIXON1989} (Continuous, Differentiable, Non-Separable, Scalable, Unimodal)
\begin{eqnarray}
    f_{48}(\textbf{x}) &=& (x_{1} - 1)^2 + \sum_{i=2}^{D}i(2x_{i}^2 - x_{i-1})^2 \nonumber
\end{eqnarray}
subject to  $-10 \leq x_{i} \leq 10$.  The global minimum is located at $\textbf{x}^{*} = f(2^(\frac{2^i-2}{2^i}))$,
$f(\textbf{x}^{*}) = 0$.  \\

\item \textbf{Dolan  Function} (Continuous, Differentiable, Non-Separable, Non-Scalable, Multimodal)
\begin{eqnarray}
	       f_{49}(\textbf{x}) &=& (x_{1} + 1.7x_{2})\sin(x_{1}) - 1.5x_{3} - 0.1x_{4}\cos(x_{4} + x_{5} - x_{1}) + \nonumber \\
                              & & 0.2x_{5}^{2} - x_{2} - 1 \nonumber
\end{eqnarray}
subject to  $-100 \leq x_{i} \leq 100$.  The global minimum is $f(\textbf{x}^{*}) = 0 $. \\

\item \textbf{Easom Function} \cite{CHUNG1998}(Continuous, Differentiable, Separable, Non-Scalable, Multimodal)
\begin{eqnarray}
	       f_{50}(\textbf{x}) &=& -\textrm{cos}(x_{1})\textrm{cos}(x_{2})\exp[-(x_{1} - \pi)^{2} \nonumber \\
                                           & & - (x_{2} - \pi)^{2}] \nonumber
\end{eqnarray}
    subject to  $-100 \leq x_{i} \leq 100$.  The global minimum is located at $\textbf{x}^{*} = f(\pi, \pi)$, $f(\textbf{x}^{*}) = -1 $. \\

\item \textbf{El-Attar-Vidyasagar-Dutta Function} \cite{EL-ATTAR1979} (Continuous, Differentiable, Non-Separable, Non-Scalable, Unimodal)
\begin{eqnarray}
	       f_{51}(\textbf{x}) &=& (x_{1}^{2} + x_{2} - 10)^2 + (x_{1} + x_{2}^{2} - 7)^2  + \nonumber\\
                              & & (x_{1}^{2} + x_{2}^{3} - 1)^2 \nonumber
\end{eqnarray}
subject to  $-500 \leq x_{i} \leq 500$.  The global minimum is located at $\textbf{x}^{*} = f(2.842503, 1.920175)$,
$f(\textbf{x}^{*}) = 0.470427 $. \\

\item \textbf{Egg Crate Function} (Continuous, Separable, Non-Scalable)
\begin{eqnarray}
    f_{52}(\textbf{x}) &=& x_{1}^{2} + x_{2}^{2} + 25(\textrm{sin}^{2}(x_{1}) + \textrm{sin}^{2}(x_{2})) \nonumber
\end{eqnarray}
    subject to  $-5 \leq x_{i} \leq 5$.  The global minimum is located at $\textbf{x}^{*} = f(0, 0)$,
$f(\textbf{x}^{*}) = 0$.\\

\item \textbf{Egg Holder Function} (Continuous, Differentiable, Non-Separable, Scalable, Multimodal)
\begin{eqnarray}
	       f_{53}(\textbf{x}) &=& \sum_{i=1}^{m-1} [-(x_{i+1} + 47)\textrm{sin}\sqrt{|x_{i+1} + x_{i}/2 + 47|}  \nonumber \\
                                           & & - x_{i}\textrm{sin}\sqrt{|x_{i} -( x_{i+1} + 47)|} ] \nonumber
        \end{eqnarray}
    subject to  $-512 \leq x_{i} \leq 512$.  The global minimum is located at $\textbf{x}^{*} = f(512, 404.2319)$,
    $f(\textbf{x}^{*}) \approx 959.64 $.\\

\item \textbf{Exponential Function} \cite{RAHNAMAYAN2007_1} (Continuous, Differentiable, Non-Separable, Scalable, Multimodal)
\begin{eqnarray}
     f_{54}(\textbf{x}) &=&  - \exp \left( { - 0.5\sum\limits_{i = 1}^D {x_i^2} } \right) \nonumber	
\end{eqnarray}
    subject to  $-1 \leq x_{i} \leq 1$.  The global minima is located at $\textbf{x} = f(0,\cdots,0)$,
    $f(\textbf{x}^{*}) = 1$.  \\

\item \textbf{Exp 2 Function} \cite{ADORIO2005} (Separable)
    \begin{eqnarray}
        f_{55}(\textbf{x}) &=& \sum_{i=0}^9\left( e^{-i x_1 / 10} - 5 e^{-i x_2 / 10} -  e^{-i / 10} \nonumber
        + 5  e^{-i} \right)^2
        \end{eqnarray}
    with domain $0 \leq x_{i} \le 20$.   The global minimum is located at $\textbf{x}^{*} = f(1,10)$,
    $f(\textbf{x}^{*}) = 0$.\\

\item \textbf{Freudenstein Roth Function} \cite{RAO2009} (Continuous, Differentiable, Non-Separable, Non-Scalable, Multimodal)
\begin{eqnarray}
    f_{56}(\textbf{x}) &=& (x_{1} - 13 + ((5-x_{2})x_{2}-2)x_{2})^{2} + \nonumber \\
                                           & & (x_{1} - 29 + ((x_{2} + 1)x_{2} - 14)x_{2})^{2} \nonumber
\end{eqnarray}
    subject to  $-10 \leq x_{i} \leq 10$.  The global minimum is located at $\textbf{x}^{*} = f(5,4)$,
    $f(\textbf{x}^{*}) = 0$. \\

\item \textbf{Giunta Function} \cite{MISHRA2006_6} (Continuous, Differentiable, Separable, Scalable, Multimodal)
\begin{eqnarray}
	       f_{57}(\textbf{x}) &=& 0.6 + \sum_{i=1}^{2}[\textrm{sin}(\frac{16}{15}x_{i} - 1)  \nonumber \\
                                          & &  + \textrm{sin}^{2}(\frac{16}{15}x_{i} - 1) \nonumber \\
                                          & &  + \frac{1}{50}\textrm{sin}(4(\frac{16}{15}x_{i} - 1))] \nonumber
\end{eqnarray}
subject to  $-1 \leq x_{i} \leq 1$.  The global minimum is located at $\textbf{x}^{*} = f(0.45834282,0.45834282)$,
$f(\textbf{x}^{*}) = 0.060447 $. \\

\item \textbf{Goldstein Price Function} \cite{GOLDSTEIN1971} (Continuous, Differentiable, Non-separable, Non-Scalable, Multimodal)
\begin{eqnarray}
    f_{58}(\textbf{x}) &=& [1  + (x_{1} + x_{2} + 1)^{2}(19 - 14x_{1}  \nonumber \\
    & & +  3x_{1}^{2} - 14x_{2} +  6x_{1}x_{2} +  3x_{2}^{2})] \nonumber \\
    & & \times [30 + (2x_{1} - 3x_{2})^{2} \nonumber \\
    & & (18 - 32x_{1} + 12x_{1}^{2} + 48x_{2} - 36x_{1}x_{2} + 27x_{2}^{2})] \nonumber
\end{eqnarray}
subject to  $-2 \leq x_{i} \leq 2$.  The global minimum is located at $\textbf{x}^{*} = f(0,-1)$,
$f(\textbf{x}^{*}) = 3 $. \\

\item \textbf{Griewank Function} \cite{GRIEWANK1981} (Continuous, Differentiable, Non-Separable, Scalable, Multimodal)
    \begin{eqnarray}
	       f_{59}(\textbf{x}) &=& \sum_{i=1}^{n}\frac{x_{i}^{2}}{4000} - \prod\textrm{cos}(\frac{x_{i}}{\sqrt{i}}) + 1 \nonumber
    \end{eqnarray}
    subject to  $-100 \leq x_{i} \leq 100$.  The global minima is located at $\textbf{x}^{*} = f(0,\cdots,0)$,
    $f(\textbf{x}^{*}) = 0$. \\

\item \textbf{Gulf Research Problem} \cite{SHANNO1970} (Continuous, Differentiable, Non-Separable, Non-Scalable, Multimodal)
\begin{eqnarray}
     f_{60}(\textbf{x}) = {\sum\limits_{i = 1}^{99} {\left[ {\exp \left( { - \frac{{{{\left( {{u_i} - {x_2}} \right)}^{{x_3}}}}}{{{x_i}}}} \right) - 0.01i} \right]} ^2} \nonumber
\end{eqnarray}

where $u_{i}=25 +[-50\ln(0.01i)]^{1/1.5}$ subject to $0.1 \leq x_{1} \leq 100$, $0 \leq x_{2} \leq 25.6$ and $0 \leq x_{1} \leq 5$.  The global minimum is located at
$\textbf{x}^{*} = f(50, 25, 1.5)$, $f(\textbf{x}^{*}) = 0$. \\

\item \textbf{Hansen Function} \cite{FRALEY1989} (Continuous, Differentiable, Separable, Non-Scalable, Multimodal)
\begin{eqnarray}
    f_{61}(\textbf{x}) &=& \sum_{i}^{4}(i + 1)\textrm{cos}(ix_{1} + i + 1) \nonumber \\
                                           & & \sum_{j=0}^{4}(j+1)\textrm{cos}((j + 2)x_{2} + j + 1) \nonumber
\end{eqnarray}
    subject to  $-10 \leq x_{i} \leq 10$.  The multiple global minima are located at
    \[\textbf{x}^{*} = f( \{-7.589893, -7.708314\}, ~~\{-7.589893, -1.425128\},\]
                                                                                     \[ \{-7.589893, ~~4.858057\}, ~~\{-1.306708, -7.708314\}, \]
                                                                                     \[ \{-1.306708, ~~4.858057\}, ~~\{~~4.976478,  ~~4.858057\},\]                                                                                     \[ \{~~4.976478, -1.425128\},  ~~\{~~4.976478, -7.708314\}), \]
\item \textbf{Hartman 3 Function} \cite{HARTMAN1972} (Continuous, Differentiable, Non-Separable, Non-Scalable, Multimodal)
\begin{eqnarray}
	       f_{62}(\textbf{x}) &=&   - \sum\limits_{i = 1}^4 {{c_i}\exp \left[ { - \sum\limits_{j = 1}^3 {{a_{ij}}{{\left( {{x_j} - {p_{ij}}} \right)}^2}} } \right]} \nonumber
\end{eqnarray}
subject to  $0 \leq x_{j} \leq 1$, $j \in \{1,2,3\}$ with constants $a_{ij}$, $p_{ij}$ and $c_{i}$ are given as \\


$\textbf{A} = [A_{ij}] = \left( {\begin{matrix}
3&{10}&{30}\\
{0.1}&{10}&{35}\\
3&{10}&{30}\\
{0.1}&{10}&{35}
\end{matrix}} \right)$, $\textbf{\underline{c}} = {c_i} = \left[ {\begin{matrix}
1\\
{1.2}\\
3\\
{3.2}
\end{matrix}} \right]$, \\
$\textbf{\underline{p}} = {p_i} = \left( {\begin{matrix}
{0.3689}&{0.1170}&{0.2673}\\
{0.4699}&{0.4837}&{0.7470}\\
{0.1091}&{0.8732}&{0.5547}\\
{0.03815}&{0.5743}&{0.8828}
\end{matrix}} \right)$  \\

The global minimum is located at $\textbf{x}^{*} = f(0.1140, 0.556, 0.852)$,
$f(\textbf{x}^{*})\approx -3.862782 $. \\

\item \textbf{Hartman 6 Function} \cite{HARTMAN1972} (Continuous, Differentiable, Non-Separable, Non-Scalable, Multimodal)
\begin{eqnarray}
	       f_{63}(\textbf{x}) &=&  - \sum\limits_{i = 1}^4 {{c_i}\exp \left[ { - \sum\limits_{j = 1}^6 {{a_{ij}}{{\left( {{x_j} - {p_{ij}}} \right)}^2}} } \right]} \nonumber
\end{eqnarray}
subject to  $0 \leq x_{j} \leq 1$, $j \in \{1,\cdots,6\}$ with constants $a_{ij}$, $p_{ij}$ and $c_{i}$ are given as \\

$\textbf{A} = [A_{ij}] = \left( {\begin{matrix}
{10}&3&{17}&{3.5}&{1.7}&8\\
{0.05}&{10}&{17}&{0.1}&8&{14}\\
3&{3.5}&{1.7}&{10}&{17}&8\\
{17}&8&{0.05}&{10}&{0.1}&{14}
\end{matrix}} \right)$, $\textbf{\underline{c}} = {c_i} = \left[ {\begin{matrix}
1\\
{1.2}\\
3\\
{3.2}
\end{matrix}} \right]$  \\
$\textbf{\underline{p}} = {p_i} = \left( {\begin{matrix}
\small{0.1312}&{0.1696}&{0.5569}&{0.0124}&{0.8283}&{0.5586}\\
{0.2329}&{0.4135}&{0.8307}&{0.3736}&{0.1004}&{0.9991}\\
{0.2348}&{0.1451}&{0.3522}&{0.2883}&{0.3047}&{0.6650}\\
{0.4047}&{0.8828}&{0.8732}&{0.5743}&{0.1091}&{0.0381}
\end{matrix}} \right)$ \\

The global minima is located at
$\textbf{x} = f(0.201690, 0.150011, 0.476874,0.275332, ...\\
0.311652, 0.657301)$,
$f(\textbf{x}^{*}) \approx -3.32236 $. \\
\item \textbf{Helical Valley} \cite{FLETCHER1963} (Continuous, Differentiable, Non-Separable, Scalable, Multimodal)
\begin{eqnarray}
    f_{64}(\textbf{x}) &=& 100\left[ {{{({x_2} - 10\theta )}^2} + \left( {\sqrt {x_1^2 + x_2^2}  - 1} \right)} \right] \nonumber \\
                                    & & + x_3^2 \nonumber
    \end{eqnarray}
where
\begin{eqnarray}
    \theta  = \left\{ \begin{array}{l}
\frac{1}{{2\pi }}{\tan ^{ - 1}}\left( {\frac{{{x_1}}}{{{x_2}}}} \right),{\rm{~ ~ ~ ~ ~ ~ ~ ~ ~ ~if }}~{x_1} \ge 0 \nonumber\\
\frac{1}{{2\pi }}{\tan ^{ - 1}}\left( {\frac{{{x_1}}}{{{x_2}}} + 0.5} \right){\rm{~ ~ ~ ~ if }}~{x_1} < 0
\end{array} \right.
\end{eqnarray}
subject to  $-10 \leq x_{i} \leq 10$.  The global minima is located at $\textbf{x}^{*} = f(1,0,0)$,  $f(\textbf{x}^{*}) = 0 $. \\

\item \textbf{Himmelblau Function} \cite{HIMMELBLAU1972} (Continuous, Differentiable, Non-Separable, Non-Scalable, Multimodal)
    \begin{eqnarray}
        f_{65}(\textbf{x}) &=& (x_{1}^{2} + x_{2} - 11)^{2} + (x_{1} + x_{2}^{2} - 7)^{2}    \nonumber
    \end{eqnarray}
    subject to  $-5 \leq x_{i} \leq 5$.  The global minimum is located at $\textbf{x}^{*} = f(3,2)$, $f(\textbf{x}^{*}) = 0 $.  \\

\item \textbf{Hosaki Function} \cite{BEKEY1974} (Continuous, Differentiable, Non-Separable, Non-Scalable, Multimodal)
\begin{eqnarray}
        f_{66}(\textbf{x}) &=& (1 - 8x_{1} + 7x_{1}^2-7/3x_{1}^3 + 1/4x_{1}^4)x_{2}^2e^{-x_{2}} \nonumber
    \end{eqnarray}
    subject to  $0 \leq x_{1} \leq 5$ and $0 \leq x_{2} \leq 6$.
    The global minimum is located at $\textbf{x}^{*} = f(4,2)$, $f(\textbf{x}^{*}) \approx -2.3458 $. \\

\item \textbf{Jennrich-Sampson Function} \cite{JENNRICH1968} (Continuous, Differentiable, Non-Separable, Non-Scalable, Multimodal)
\begin{eqnarray}
     f_{67}(\textbf{x}) &=& {\sum\limits_{i = 1}^{10} {\left( {2 + 2i - \left( {{{\mathop{\rm e}\nolimits} ^{i{x_1}}} + {e^{i{x_2}}}} \right)} \right)} ^2} \nonumber
\end{eqnarray}
subject to $-1 \leq x_{i} \leq 1$.  The global minimum is located at $\textbf{x}^{*} = f(0.257825,0.257825)$, $f(\textbf{x}^{*}) = 124.3612 $.  \\

\item \textbf{Langerman-5 Function} \cite{BERSINI1996} (Continuous, Differentiable, Non-Separable, Scalable, Multimodal)
\begin{eqnarray}
    f_{68}(\textbf{x}) &=& -\sum_{i=1}^{m}c_{i}e^{-\frac{1}{\pi} \sum_{j=1}^{D}(x_{j} - a_{ij})^2}\textrm{cos}\left(\pi \sum_{j=1}^{D}(x_{j} - a_{ij})^2\right) \nonumber
\end{eqnarray}

subject to $ 0 \le x_j \le 10$, where $j \in [0, D-1]$ and $m=5$. It has a global minimum value of $f(\textbf{x}^{*}) = -1.4$.
The matrix $A$ and column vector $c$ are given as

The matrix $A$ is given by
{\small
$$
\textbf{{A}}=[A_{ij}] =\left[
\begin{array}{cccccccccc}
    9.681& 0.667&4.783& 9.095& 3.517& 9.325& 6.544& 0.211& 5.122& 2.020  \\
    9.400& 2.041& 3.788& 7.931& 2.882& 2.672& 3.568& 1.284& 7.033& 7.374\\
    8.025& 9.152& 5.114& 7.621& 4.564& 4.711& 2.996& 6.126& 0.734& 4.982\\
    2.196& 0.415& 5.649& 6.979& 9.510& 9.166& 6.304& 6.054& 9.377& 1.426\\
    8.074& 8.777& 3.467& 1.863& 6.708& 6.349& 4.534& 0.276& 7.633& 1.567\\
\end{array}
\right]
$$
}
$\textbf{\underline{c}}= {c_i} = \left[ {\begin{array}{rrrrr}
{0.806}\\
{0.517}\\
{1.5}\\
{0.908}\\
{0.965}
\end{array}} \right]$

\item \textbf{Keane Function} (Continuous, Differentiable, Non-Separable, Non-Scalable, Multimodal)
    \begin{eqnarray}
        f_{69}(\textbf{x}) &=& \frac{\textrm{sin}^{2}(x_{1} - x_{2})\textrm{sin}^{2}(x_{1} + x_{2})}{\sqrt{x_{1}^{2} + x_{2}^{2}}}    \nonumber
    \end{eqnarray}
    subject to  $0 \leq x_{i} \leq 10$.

    The multiple global minima are located at $\textbf{x}^{*}$ =
    $f(\{0, 1.39325\}$,$\{1.39325, 0\})$, $f(\textbf{x}^{*})$ =$ -0.673668$. \\

\item \textbf{Leon Function} \cite{LAVI1966}(Continuous, Differentiable, Non-Separable, Non-Scalable, Unimodal)
    \begin{eqnarray}
        f_{70}(\textbf{x}) &=& 100(x_{2} - x_{1}^{2})^2 + (1 - x_{1})^{2} \nonumber
    \end{eqnarray}
    subject to  $-1.2 \leq x_{i} \leq 1.2$.  A global minimum is located at $f(\textbf{x}^{*}) = f(1,1)$, $f(\textbf{x}^{*}) = 0$.\\

\item \textbf{Matyas Function} \cite{HEDAR} (Continuous, Differentiable, Non-Separable, Non-Scalable, Unimodal)
    \begin{eqnarray}
        f_{71}(\textbf{x}) &=& 0.26(x_{1}^2 + x_{2}^2) - 0.48x_{1}x_{2} \nonumber
    \end{eqnarray}
    subject to  $-10 \leq x_{i} \leq 10$.  The global minimum is located at $\textbf{x}^{*} = f(0,0)$, $f(\textbf{x}^{*}) = 0 $. \\

\item \textbf{McCormick Function} \cite{LOOTSMA1972} (Continuous, Differentiable, Non-Separable, Non-Scalable, Multimodal)
\begin{eqnarray}
    f_{72}(\textbf{x}) &=& \sin(x_{1} + x_{2}) + (x_{1} - x_{2})^2 - (3/2)x_{1} + (5/2)x_{2} + 1 \nonumber
\end{eqnarray}
subject to $-1.5 \leq x_{1} \leq 4$ and $-3 \leq x_{2} \leq 3$.  The global minimum is located at $\textbf{x}^{*} = f(-0.547,-1.547)$,
$f(\textbf{x}^{*}) \approx -1.9133 $. \\

\item \textbf{Miele Cantrell Function} \cite{CRAGG1969} (Continuous, Differentiable, Non-Separable, Non-Scalable, Multimodal)
\begin{eqnarray}
        f_{73}(\textbf{x}) &=& {\left( {{e^{ - {x_1}}} - {x_2}} \right)^4} + 100{\left( {{x_2} - {x_3}} \right)^6} \nonumber \\
                                        & & + {\left( {\tan \left( {{x_3} - {x_4}} \right)} \right)^4} + x_1^8 \nonumber
\end{eqnarray}
subject to  $-1 \leq x_{i} \leq 1 $.  The global minimum is located at $\textbf{x}^{*} = f(0,1,1,1)$, $f(\textbf{x}^{*}) = 0$. \\

\item \textbf{Mishra 1 Function} \cite{MISHRA2006_1} (Continuous, Differentiable, Non-Separable, Scalable, Multimodal)
\begin{eqnarray}
        f_{74}(\textbf{x}) &=& \left(1 + D - \sum_{i=1}^{N-1}x_{i}\right)^{N - \sum_{i=1}^{N-1}x_{i}}
\end{eqnarray}
subject to  $0 \leq x_{i} \leq 1 $.  The global minimum is $f(\textbf{x}^{*}) = 2$. \\

\item \textbf{Mishra 2 Function} \cite{MISHRA2006_1} (Continuous, Differentiable, Non-Separable, Scalable, Multimodal)
\begin{eqnarray}
        f_{75}(\textbf{x}) &=& \left(1 + D - \sum_{i=1}^{N-1}0.5(x_{i} + x_{i+1})\right)^{N - \sum_{i=1}^{N-1}0.5(x_{i} + x_{i+1})}
\end{eqnarray}
subject to  $0 \leq x_{i} \leq 1 $.  The global minimum is $f(\textbf{x}^{*}) = 2$. \\

\item \textbf{Mishra 3 Function} \cite{MISHRA2006_6} (Continuous, Differentiable, Non-Separable, Non-Scalable, Multimodal)
\begin{eqnarray}
        f_{76}(\textbf{x}) &=& \sqrt{\bigl\lvert {\cos{\sqrt{\bigl \lvert x_{1}^2 + x_{2}^2 \bigr\rvert}}\bigr \rvert}} + 0.01(x_{1} + x_{2}) \nonumber
\end{eqnarray}
The global minimum is located at $\textbf{x}^{*} = f(-8.466, -10)$, $f(\textbf{x}^{*}) = -0.18467$. \\

\item \textbf{Mishra 4 Function} \cite{MISHRA2006_6} (Continuous, Differentiable, Non-Separable, Non-Scalable, Multimodal)
\begin{eqnarray}
        f_{77}(\textbf{x}) &=& \sqrt{\bigl \lvert {\sin{\sqrt{\bigl \lvert x_{1}^2 + x_{2}^2 \bigr \rvert }\bigr \rvert}}} + 0.01(x_{1} + x_{2}) \nonumber
\end{eqnarray}
The global minimum is located at $\textbf{x}^{*} = f(-9.94112, -10)$, $f(\textbf{x}^{*}) = -0.199409$. \\

\item \textbf{Mishra 5 Function} \cite{MISHRA2006_6} (Continuous, Differentiable, Non-Separable, Non-Scalable, Multimodal)
\begin{eqnarray}
        f_{78}(\textbf{x}) &=& \Big[\sin^2(\cos((x_1) + \cos(x_2)))^2 + \cos^2(\sin(x_1) + \sin(x_2)) + x_{1} \Big]^2 \nonumber \\
                           & & + 0.01(x_{1} + x_{2}) \nonumber
\end{eqnarray}
The global minimum is located at $\textbf{x}^{*} = f(-1.98682, -10)$, $f(\textbf{x}^{*}) = -1.01983$. \\

\item \textbf{Mishra 6 Function} \cite{MISHRA2006_6} (Continuous, Differentiable, Non-Separable, Non-Scalable, Multimodal)
\begin{eqnarray}
        f_{79}(\textbf{x}) &=& -\ln\Big[\sin^2(\cos((x_1) + \cos(x_2)))^2 - \cos^2(\sin(x_1) + \sin(x_2)) + x_{1} \Big]^2 \nonumber \\
                           & & + 0.01((x_{1} - 1)^2 + (x_{2} - 1)^2) \nonumber
\end{eqnarray}
The global minimum is located at $\textbf{x}^{*} = f(2.88631, 1.82326)$, $f(\textbf{x}^{*}) = -2.28395$. \\

\item \textbf{Mishra 7 Function} (Continuous, Differentiable, Non-Separable, Non-Scalable, Multimodal)
\begin{eqnarray}
        f_{80}(\textbf{x}) &=& \Big[\prod_{i=1}^{D}x_{i} - N!\Big]^2 \nonumber
\end{eqnarray}
The global minimum is $f(\textbf{x}^{*}) = 0$. \\

\item \textbf{Mishra 8 Function} \cite{MISHRA2006_6} (Continuous, Differentiable, Non-Separable, Non-Scalable, Multimodal)
\begin{eqnarray}
   f_{81}(\textbf{x}) &=& 0.001\Big[\Big|x_{1}^{10} - 20x_{1}^{9} + 180x_{1}^{8} - 960x_{1}^{7} +3360x_{1}^{6} - 8064x_{1}^{5} \nonumber\\
                      & & 1334x_{1}^4 - 15360x_{1}^3 + 11520x_{1}^2 -5120x_{1} +2624\Big| \nonumber \\
                      & & \Big| x_{2}^4 + 12x_{2}^3 + 54x_{2}^2 +108x_{2} + 81\Big|\Big]^2 \nonumber
\end{eqnarray}
The global minimum is located at $\textbf{x}^{*} = f(2, -3)$, $f(\textbf{x}^{*}) = 0$. \\

\item \textbf{Mishra 9 Function} \cite{MISHRA2006_6} (Continuous, Differentiable, Non-Separable, Non-Scalable, Multimodal)
\begin{eqnarray}
   f_{82}(\textbf{x}) &=& \Big[ab^2c + abc^2 + b^2 + (x_1 + x_2 - x_3)^2\Big]^2 \nonumber
\end{eqnarray}
where $ a = 2x_{1}^{3} + 5x_1x_2 + 4x_3 -2x_1^2x_3 -18 $, $b = x_{1} + x_2^3 + x_1x_3^2 - 22 $ \\
$ c = 8x_1^2 + 2x_2x_3 + 2x_2^2 + 3x_2^3 - 52$.
The global minimum is located at $\textbf{x}^{*} = f(1, 2, 3)$, $f(\textbf{x}^{*}) = 0$. \\

\item \textbf{Mishra 10 Function} \cite{MISHRA2006_6} (Continuous, Differentiable, Non-Separable, Non-Scalable, Multimodal)
\begin{eqnarray}
   f_{83}(\textbf{x}) &=& \Big[\lfloor x_{1} \perp x_{2}\rfloor - \lfloor x_{1} \rfloor - \lfloor x_{2} \rfloor\Big]^2 \nonumber
\end{eqnarray}
The global minimum is located at $\textbf{x}^{*} = f\{(0, 0), (2, 2)\}$, $f(\textbf{x}^{*}) = 0$. \\

\item \textbf{Mishra 11 Function} \cite{MISHRA2006_6} (Continuous, Differentiable, Non-Separable, Non-Scalable, Multimodal)
\begin{eqnarray}
   f_{84}(\textbf{x}) &=& \Big[\frac{1}{D}\sum_{i=1}^{D}\big|x_i\big| - \big(\prod_{i=1}^{D}\big|x_i\big|\big)^{\frac{1}{N}}\Big]^2 \nonumber
\end{eqnarray}
The global minimum is $f(\textbf{x}^{*}) = 0$. \\

\item \textbf{Parsopoulos Function} (Continuous, Differentiable, Separable, Scalable, Multimodal)
\begin{eqnarray}
    f_{85}(\textbf{x}) &=& \cos {\left( {{x_1}} \right)^2} + \sin {\left( {{x_2}} \right)^2} \nonumber
\end{eqnarray}
subject to $-5 \leq x_{i} \leq 5$, where $(x_{1},x_{2}) \in \mathbb{R}^{2}$.  This function has infinite number of global minima in $\mathbb{R}^{2}$, at points
$(\kappa\frac{\pi}{2}, \lambda \pi)$, where $\kappa = \pm 1,\pm 3,...$ and $\lambda = 0,\pm 1, \pm 2, ...$.  In the given domain problem, function has
12 global minima all equal to zero.

\item \textbf{Pen Holder Function} \cite{MISHRA2006_6}  (Continuous, Differentiable, Non-Separable, Non-Scalable, Multimodal)
        \begin{eqnarray}
	       f_{86}(\textbf{x}) &=& -\exp[|\textrm{cos}(x_{1})\textrm{cos}(x_{2})
                                            e^{|1 - [(x_{1}^{2} + x_{2}^{2})]^{0.5}/\pi|}|^{-1}] \nonumber
        \end{eqnarray}
    subject to  $-11 \leq x_{i} \leq 11$.  The four global minima are located at $\textbf{x}^{*}$ = $f(\pm 9.646168$, $\pm 9.646168)$,
    $f(\textbf{x}^{*})= -0.96354 $. \\

\item \textbf{Pathological Function} \cite{RAHNAMAYAN2007} (Continuous, Differentiable, Non-Separable, Non-Scalable, Multimodal)
\begin{eqnarray}
    f_{87}(\textbf{x}) &=& \sum_{i=1}^{D-1}\left(0.5 + \frac{\sin^2\sqrt{100x_{i}^2 + x_{i+1}^2} - 0.5}{1 + 0.001(x_{i}^2 -2x_{i}x_{i+1}+x_{i+1}^2)^2}\right) \nonumber
\end{eqnarray}
subject to $-100 \leq x_{i} \leq 100$.  The global minima is located $\textbf{x}^{*} = f(0,\cdots,0)$, $f(\textbf{x}^{*}) = 0$. \\

 \item \textbf{Paviani Function} \cite{HIMMELBLAU1972} (Continuous, Differentiable, Non-Separable, Scalable, Multimodal)
 \begin{eqnarray}
    f_{88}(\textbf{x}) & = &\sum\limits_{i = 1}^{10} {\left[ {{{\left( {\ln \left( {{x_i} - 2} \right)} \right)}^2} + {{\left( {\ln \left( {10 - {x_i}} \right)} \right)}^2}} \right]}  - {\left( {\prod\limits_{i = 1}^{10} {{x_i}} } \right)^{0.2}} \nonumber
\end{eqnarray}
subject to  $2.0001 \leq x_{i} \leq 10$, $i \in {1,2,...,10}$.  The global minimum is located at $\textbf{x}^{*} \approx f(9.351,....,9.351)$, $f(\textbf{x}^{*}) \approx -45.778 $. \\

\item \textbf{Pint\'{e}r Function} \cite{PINTER1996} (Continuous, Differentiable, Non-separable, Scalable, Multimodal)
\begin{eqnarray}
	       f_{89}(\textbf{x}) &=& \sum\limits_{i = 1}^D {ix_i^2 + \sum\limits_{i = 1}^D {20i{{\sin }^2}A + } } \sum\limits_{i = 1}^D {i{{\log }_{10}}\left( {1 + i{B}^2} \right)} \nonumber
\end{eqnarray}
    where \\
    \begin{eqnarray}
        A &=& {\left( {{x_{i - 1}}\sin {x_i} + \sin {x_{i + 1}}} \right)} \nonumber \\
        B &=& {\left( {x_{i - 1}^2 - 2{x_i} + 3{x_{i + 1}} - \cos {x_i} + 1} \right)}\nonumber
    \end{eqnarray}
where $x_{0}= x_{D}$ and $x_{D+1} = x_{1}$, subject to  $-10 \leq x_{i} \leq 10$.
The global minima is located at $\textbf{x}^{*} = f(0,\cdots,0)$, $f(\textbf{x}^{*}) = 0 $. \\

\item \textbf{Periodic Function} \cite{ALI2005} (Separable)
\begin{eqnarray}
	       f_{90}(\textbf{x}) &=& 1 + \textrm{sin}^{2}(x_{1}) + \textrm{sin}^{2}(x_{2}) - 0.1e^{-(x_{1}^{2} + x_{2}^{2})} \nonumber
        \end{eqnarray}
    subject to  $-10 \leq x_{i} \leq 10$.  The global minimum is located at $\textbf{x}^{*} = f(0,0)$, $f(\textbf{x}^{*}) = 0.9 $. \\
\item \textbf{Powell Singular Function} \cite{POWELL1962} (Continuous, Differentiable, Non-Separable Scalable, Unimodal)
\begin{eqnarray}
    f_{91}(\textbf{x}) &=& \sum\limits_{i = 1}^{D/4} {{\left( {{x_{4i - 3}} + 10{x_{4i - 2}}} \right)}^2} \nonumber \\
                                & & + 5{{\left( {{x_{4i - 1}} - {x_{4i}}} \right)}^2} + {{\left( {{x_{4i - 2}} - {x_{4i - 1}}} \right)}^4} \nonumber \\
                                & & + 10{{\left( {{x_{4i - 3}} - {x_{4i}}} \right)}^4} \nonumber
\end{eqnarray}
subject to  $-4 \leq x_{i} \leq 5$.  The global minima is located at $\textbf{x}^{*} = f(3,-1,0,1,\cdots,3,-1,0,1)$, $f(\textbf{x}^{*}) = 0 $. \\

\item \textbf{Powell Singular 2 Function} \cite{FU2006} (Continuous, Differentiable, Non-Separable Scalable, Unimodal)
\begin{eqnarray}
    f_{92}(\textbf{x}) &=& \sum\limits_{i = 1}^{D-2} {{\left( {{x_{i - 1}} + 10{x_{i}}} \right)}^2} \nonumber \\
                                & & + 5{{\left( {{x_{i+1}} - {x_{i+2}}} \right)}^2} + {{\left( {{x_{i}} - {2x_{i + 1}}} \right)}^4} \nonumber \\
                                & & + 10{{\left( {{x_{i - 1}} - {x_{i+2}}} \right)}^4} \nonumber
\end{eqnarray}
subject to  $-4 \leq x_{i} \leq 5$.  The global minimum is $f(\textbf{x}^{*}) = 0 $.

\item \textbf{Powell Sum Function} \cite{RAHNAMAYAN2007} (Continuous, Differentiable, Separable Scalable, Unimodal)
\begin{eqnarray}
     f_{93}(\textbf{x}) &=& \sum_{i=1}^{D}\Big|x_{i}\Big|^{i+1} \nonumber
\end{eqnarray}
subject to  $-1 \leq x_{i} \leq 1$.  The global minimum is $f(\textbf{x}^{*}) = 0 $.

\item \textbf{Price 1 Function} \cite{PRICE1977} (Continuous, Non-Differentiable, Separable Non-Scalable, Multimodal)
\begin{eqnarray}
    f_{94}(\textbf{x}) &=& (|x_1| - 5)^2 + (|x_2| - 5)^2 \nonumber
\end{eqnarray}
subject to  $-500 \leq x_{i} \leq 500$.  The global minimum are located at
$\textbf{x}^{*}$ =$ f(\{-5, -5\}$,$\{-5, 5\}$, $\{5, -5\}$, $\{5, 5\})$, $f(\textbf{x}^{*}) = 0$. \\

\item \textbf{Price 2 Function} \cite{PRICE1977} (Continuous, Differentiable, Non-Separable Non-Scalable, Multimodal)
\begin{eqnarray}
    f_{95}(\textbf{x}) &=& 1 + \sin^2x_1 + \sin^2x_2 - 0.1e^{-x_1^2 - x_2^2} \nonumber
\end{eqnarray}
subject to  $-10 \leq x_{i} \leq 10$.  The global minimum is located at
$\textbf{x}^{*} = f(0 \cdots 0)$, $f(\textbf{x}^{*}) = 0.9$. \\

\item \textbf{Price 3 Function} \cite{PRICE1977} (Continuous, Differentiable, Non-Separable Non-Scalable, Multimodal)
\begin{eqnarray}
    f_{96}(\textbf{x}) &=& 100(x_2 - x_1^2)^2 + 6\Big[6.4(x_2 - 0.5)^2 - x_1 - 0.6 \Big]^2 \nonumber
\end{eqnarray}
subject to  $-500 \leq x_{i} \leq 500$.  The global minimum are located at
$\textbf{x}^{*}$ =$ f(\{-5, -5\}$,$\{-5, 5\}$, $\{5, -5\}$, $\{5, 5\})$, $f(\textbf{x}^{*}) = 0$. \\

\item \textbf{Price 4 Function} \cite{PRICE1977} (Continuous, Differentiable, Non-Separable Non-Scalable, Multimodal)
\begin{eqnarray}
    f_{97}(\textbf{x}) &=& (2 x_1^3 x_2 - x_2^3)^2 + (6 x_1 - x_2^2 + x_2) ^ 2 \nonumber
\end{eqnarray}
subject to $-500 \leq x_i \leq 500$.  The three global minima are located at $\textbf{x}^{*}$ =
$f(\{0, 0\}$,$\{2,4\}$, $\{1.464, -2.506\})$,
$f(\textbf{x}^{*}) = 0 $.

\item \textbf{Qing Function} \cite{QING2006} (Continuous, Differentiable, Separable Scalable, Multimodal)
\begin{eqnarray}
    f_{98}(\textbf{x}) &=& \sum_{i=1}^{D}(x_i^2 - i)^2 \nonumber
\end{eqnarray}
subject to $-500 \leq x_i \leq 500$.  The global minima are located at $\textbf{x}^{*} = f(\pm \sqrt{i})$,
$f(\textbf{x}^{*}) = 0 $.

\item \textbf{Quadratic Function} (Continuous, Differentiable, Non-Separable, Non-Scalable)
\begin{eqnarray}
	 f_{99}(\textbf{x}) &=& -3803.84 - 138.08x_{1} - 232.92x_{2} \nonumber \\
                                           & & + 128.08x_{1}^{2} \nonumber + 203.64x_{2}^{2} + 182.25x_{1}x_{2} \nonumber
        \end{eqnarray}
    subject to  $-10 \leq x_{i} \leq 10$.  The global minimum is located at $\textbf{x}^{*} = f(0.19388, 0.48513)$, $f(\textbf{x}^{*}) = -3873.7243 $. \\

\item \textbf{Quartic Function} \cite{STORN1996} (Continuous, Differentiable, Separable, Scalable)
\begin{eqnarray}
	       f_{100}(\textbf{x}) &=&\sum_{i=1}^{D}ix_{i}^{4} + \textrm{random}[0,1) \nonumber
\end{eqnarray}
subject to  $-1.28 \leq x_{i} \leq 1.28$.  The global minima is located at $\textbf{x}^{*} = f(0,\cdots,0)$, $f(\textbf{x}^{*}) = 0 $. \\

\item \textbf{Quintic Function} \cite{MISHRA2006_6}(Continuous, Differentiable, Separable, Non-Scalable, Multimodal)
\begin{eqnarray}
	       f_{101}(\textbf{x}) &=& \sum_{i=1}^{D}\lvert x_{i}^5 - 3x_{i}^4 + 4x_{i}^3 + 2x_{i}^2 -10x_{i} - 4\rvert \nonumber
        \end{eqnarray}
    subject to  $-10 \leq x_{i} \leq 10$.  The global minimum is located at $\textbf{x}^{*} = f($-1$ ~or~ $2$)$, $f(\textbf{x}^{*}) = 0 $. \\

\item \textbf{Rana Function} \cite{PRICE2005} (Continuous, Differentiable, Non-Separable, Scalable, Multimodal)
\begin{eqnarray}
	       f_{102}(\textbf{x}) &=& \sum_{i=0}^{D-2}(x_{i+1} + 1)\textrm{cos}(t_{2})\textrm{sin}(t_{1}) + x_{i}*\textrm{cos}(t_{1})\textrm{sin}(t_{2}) \nonumber
\end{eqnarray}
subject to $-500 \leq x_{i} \leq 500$, where $t_{1} = \sqrt{\|x_{i+1} + x_{i} + 1\|}$ and $t_{2} = \sqrt{\|x_{i+1} - x_{i} + 1\|}$.

\item \textbf{Ripple 1 Function} (Non-separable)
\begin{eqnarray}
    f_{103}(\textbf{x}) &=& \sum_{i=1}^{2} -e^{\textrm{-2 ln2}(\frac{x_{i} - 0.1}{0.8})^2}(\textrm{sin}^{6}(5\pi x_{i}) + 0.1\textrm{cos}^2(500\pi x_{i})) \nonumber
\end{eqnarray}
subject to $0 \leq x_{i} \leq 1$.  It has one global minimum and 252004 local minima. The global form of the function
consists of $25$ holes, which forms a $5 \times 5$ regular grid.  Additionally, the whole function landscape is full of small
ripples caused by high frequency cosine function which creates a large number of local minima.

\item \textbf{Ripple 25 Function} (Non-separable)
\begin{eqnarray}
    f_{104}(\textbf{x}) &=& \sum_{i=1}^{2} -e^{\textrm{-2 ln2}(\frac{x_{i} - 0.1}{0.8})^2}(\textrm{sin}^{6}(5\pi x_{i})) \nonumber
\end{eqnarray}
subject to $0 \leq x_{i} \leq 1$.  It has one global form of the Ripple-1 function without any ripples due to absence of cosine
term.

\item \textbf{Rosenbrock Function} \cite{ROSENBROCK1960} (Continuous, Differentiable, Non-Separable, Scalable, Unimodal)
\begin{eqnarray}
	       f_{105}(\textbf{x}) &=& \sum_{i=1}^{D-1}\left[100(x_{i+1} - x_{i}^{2})^2 + (x_{i} - 1)^2\right] \nonumber
\end{eqnarray}
subject to  $-30 \leq x_{i} \leq 30$.  The global minima is located at $\textbf{x}^{*} = f(1,\cdots,1)$, $f(\textbf{x}^{*}) = 0$. \\

\item \textbf{Rosenbrock Modified Function} (Continuous, Differentiable, Non-Separable, Non-Scalable, Multimodal)
\begin{eqnarray}
	       f_{106}(\textbf{x}) &=& 74 + 100(x_{2} - x_{1}^2)^2 + (1 - x)^2 \nonumber \\
                              & & - 400e^{-\frac{(x_{1} + 1)^2+(x_{2} + 1)^2}{0.1}} \nonumber
\end{eqnarray}
subject to  $-2 \leq x_{i} \leq 2$.  In this function, a Gaussian bump at $(-1,1)$ is added, which causes a local minimum at $(1,1)$ and global minimum
is located at $\textbf{x}^{*} = f(-1,-1)$, $f(\textbf{x}^{*}) = 0$.
This modification makes it a difficult to optimize because local minimum basin is larger than the global minimum basin. \\

\item \textbf{Rotated Ellipse Function} (Continuous, Differentiable, Non-Separable, Non-Scalable, Unimodal)
\begin{eqnarray}
	       f_{107}(\textbf{x}) &=& 7x_1^2 - 6\sqrt{3}x_1x_2 + 13x_2^2  \nonumber
\end{eqnarray}
subject to  $-500 \leq x_{i} \leq 500$.  The global minimum is located at $\textbf{x}^{*} = f(0, 0)$, $f(\textbf{x}^{*}) = 0$. \\

\item \textbf{Rotated Ellipse 2 Function} \cite{PRICE2005} (Continuous, Differentiable, Non-Separable, Non-Scalable, Unimodal)
\begin{eqnarray}
	       f_{108}(\textbf{x}) &=& x_1^2 - x_1x_2 + x_2^2  \nonumber
\end{eqnarray}
subject to  $-500 \leq x_{i} \leq 500$.  The global minimum is located at $\textbf{x}^{*} = f(0, 0)$, $f(\textbf{x}^{*}) = 0$. \\s

\item \textbf{Rump Function} \cite{MOORE1988} (Continuous, Differentiable, Non-Separable, Non-Scalable, Unimodal)
\begin{eqnarray}
	       f_{109}(\textbf{x}) &=& (333.75 - x_1^2)x_2^6 + x_1^2(11x_1^2x_2^2 - 121x_2^4 - 2) + 5.5x_2^8 +\frac{x_1}{2x_2} \nonumber
\end{eqnarray}
subject to  $-500 \leq x_{i} \leq 500$.  The global minimum is located at $\textbf{x}^{*} = f(0, 0)$, $f(\textbf{x}^{*}) = 0$. \\

\item \textbf{Salomon Function} \cite{SALOMON1996} (Continuous, Differentiable, Non-Separable, Scalable, Multimodal)
\begin{eqnarray}
	       f_{110}(\textbf{x}) &=& = 1 - \cos \Bigg(2\pi \sqrt{\sum_{i=1}^D x_i^2} \Bigg) + 0.1 \sqrt{\sum_{i=1}^D x_i^2} \nonumber
\end{eqnarray}
subject to $-100 \leq x_{i} \leq 100$. The global minimum is located at $\textbf{x}^{*} = f(0,0)$, $f(\textbf{x}^{*}) = 0$. \\

\item \textbf{Sargan Function} \cite{DIXON1978} (Continuous, Differentiable, Non-Separable, Scalable, Multimodal)
\begin{eqnarray}
    f_{111}(\textbf{x}) &=& = \sum_{i=1}{D}\Big(x_{i}^2 + 0.4\sum_{j\neq1}x_ix_j\Big) \nonumber
\end{eqnarray}
subject to $-100 \leq x_{i} \leq 100$. The global minimum is located at $\textbf{x}^{*} = f(0,\cdots, 0)$, $f(\textbf{x}^{*}) = 0$. \\

\item  \textbf{Scahffer 1 Function} \cite{MISHRA2006_7} (Continuous, Differentiable, Non-Separable, Non-Scalable, Unimodal)
\begin{eqnarray}
    f_{112}(\textbf{x}) &=& 0.5  + \frac{\textrm{sin}^{2}(x_{1}^{2} + x_{2}^{2})^{2} - 0.5}{1 + 0.001(x_{1}^{2} + x_{2}^{2})^{2}} \nonumber
\end{eqnarray}
subject to  $-100 \leq x_{i} \leq 100 $.  The global minimum is located at $\textbf{x}^{*} = f(0,0)$, $f(\textbf{x}^{*}) = 0$. \\

\item  \textbf{Scahffer 2 Function} \cite{MISHRA2006_7} (Continuous, Differentiable, Non-Separable, Non-Scalable, Unimodal)
\begin{eqnarray}
    f_{113}(\textbf{x}) &=& 0.5  + \frac{\textrm{sin}^{2}(x_{1}^{2} - x_{2}^{2})^{2} - 0.5}{1 + 0.001(x_{1}^{2} + x_{2}^{2})^{2}} \nonumber
\end{eqnarray}
subject to  $-100 \leq x_{i} \leq 100 $.  The global minimum is located at $\textbf{x}^{*} = f(0,0)$, $f(\textbf{x}^{*}) = 0$. \\

\item  \textbf{Scahffer 3 Function} \cite{MISHRA2006_7} (Continuous, Differentiable, Non-Separable, Non-Scalable, Unimodal)
\begin{eqnarray}
    f_{114}(\textbf{x}) &=& 0.5  + \frac{\textrm{sin}^{2}\Big(\cos\Big|x_{1}^{2} - x_{2}^{2}\Big|\Big) - 0.5}{1 + 0.001(x_{1}^{2} + x_{2}^{2})^{2}} \nonumber
\end{eqnarray}
subject to  $-100 \leq x_{i} \leq 100 $.  The global minimum is located at $\textbf{x}^{*} = f(0,1.253115)$, $f(\textbf{x}^{*}) = 0.00156685$. \\

\item  \textbf{Scahffer 4 Function} \cite{MISHRA2006_7} (Continuous, Differentiable, Non-Separable, Non-Scalable, Unimodal)
\begin{eqnarray}
    f_{115}(\textbf{x}) &=& 0.5  + \frac{\textrm{cos}^{2}\Big(\sin(x_{1}^{2} - x_{2}^{2})\Big) - 0.5}{1 + 0.001(x_{1}^{2} + x_{2}^{2})^{2}} \nonumber
\end{eqnarray}
subject to  $-100 \leq x_{i} \leq 100 $.  The global minimum is located at $\textbf{x}^{*} = f(0,1.253115)$, $f(\textbf{x}^{*}) = 0.292579$. \\

\item \textbf{Schmidt Vetters Function} \cite{LOOTSMA1972} (Continuous, Differentiable, Non-Separable, Non-Scalable, Multimodal)
\begin{eqnarray}
    f_{116}(\textbf{x}) &=&  \frac{1}{1 + (x_1 - x_2)^2} + \sin\Big(\frac{\pi x_2 + x_3}{2} \Big) \nonumber \\
                        & & + e^{({\frac{x_1 + x_2}{x_2}}-2)^2} \nonumber
\end{eqnarray}
The global minimum is located at $\textbf{x}^{*} = f(0.78547, 0.78547, 0.78547)$, $f(\textbf{x}^{*}) = 3$. \\

\item \textbf{Schumer Steiglitz Function} \cite{SCHUMER1968} (Continuous, Differentiable, Separable, Scalable, Unimodal)
\begin{eqnarray}
    f_{117}(\textbf{x}) &=&  \sum_{i=1}^{^D}{x_{i}^4} \nonumber
\end{eqnarray}
The global minimum is located at $\textbf{x}^{*} = f(0,\dots,0)$, $f(\textbf{x}^{*}) = 0$. \\

\item \textbf{Schwefel Function} \cite{SCHWEFEL1981} (Continuous, Differentiable, Partially-Separable, Scalable, Unimodal)
\begin{eqnarray}
	       f_{118}(\textbf{x}) &=&  \Big(\sum_{i=1}^{^D}{x_{i}^2}\Big)^{\alpha} \nonumber
\end{eqnarray}
where $\alpha \ge 0$, subject to  $-100 \leq x_{i} \leq 100$.  The global minima is located at $\textbf{x}^{*} = f(0,\cdots,0)$,
$f(\textbf{x}^{*}) = 0$.\\

\item \textbf{Schwefel 1.2 Function} \cite{SCHWEFEL1981}  (Continuous, Differentiable, Non-Separable, Scalable, Unimodal)
\begin{eqnarray}
	       f_{119}(\textbf{x}) &=&  \sum_{i=1}^{^D}\left(\sum_{j=1}^{i}x_{j}\right)^2 \nonumber
\end{eqnarray}
subject to  $-100 \leq x_{i} \leq 100$.  The global minima is located at $\textbf{x}^{*} = f(0,\cdots,0)$,
$f(\textbf{x}^{*}) = 0$.\\

\item \textbf{Schwefel 2.4 Function} \cite{SCHWEFEL1981} (Continuous, Differentiable, Separable, Non-Scalable, Multimodal)
\begin{eqnarray}
    f_{120}(\textbf{x}) &=&  \sum_{i=1}^{D}(x_{i} - 1)^2 + (x_1 - x_i^2)^2 \nonumber
\end{eqnarray}
subject to $0 \leq x_{i} \leq 10$.  The global minima is located at $\textbf{x}^{*} = f(1,\cdots,1)$,
$f(\textbf{x}^{*}) = 0$.\\

\item \textbf{Schwefel 2.6 Function} \cite{SCHWEFEL1981} (Continuous, Differentiable, Non-Separable, Non-Scalable, Unimodal)
\begin{eqnarray}
    f_{121}(\textbf{x}) &=&  \max (|x_1 + 2x_2 -7|,|2x_1 + x_2 - 5|) \nonumber
\end{eqnarray}
subject to $-100 \leq x_{i} \leq 100$.  The global minima is located at $\textbf{x}^{*} = f(1,3)$,
$f(\textbf{x}^{*}) = 0$.\\

\item \textbf{Schwefel 2.20 Function} \cite{SCHWEFEL1981} (Continuous, Non-Differentiable, Separable, Scalable, Unimodal)
\begin{eqnarray}
	       f_{122}(\textbf{x}) &=&  -\sum_{i=1}^{n}|x_{i}| \nonumber  
\end{eqnarray}
subject to  $-100 \leq x_{i} \leq 100$.  The global minima is located at $\textbf{x}^{*} = f(0,\cdots,0)$,
$f(\textbf{x}^{*}) = 0$.  \\

\item \textbf{Schwefel 2.21 Function} \cite{SCHWEFEL1981} (Continuous, Non-Differentiable, Separable, Scalable, Unimodal)
\begin{eqnarray}
	       f_{123}(\textbf{x}) &=&  \underset{1\le i \le D}\max|x_{i}| \nonumber  
\end{eqnarray}
subject to  $-100 \leq x_{i} \leq 100$.  The global minima is located at $\textbf{x}^{*} = f(0,\cdots,0)$,
$f(\textbf{x}^{*}) = 0$.  \\

\item \textbf{Schwefel 2.22 Function} \cite{SCHWEFEL1981} (Continuous, Differentiable, Non-Separable, Scalable, Unimodal)
\begin{eqnarray}
	       f_{124}(\textbf{x}) &=&  \sum_{i=1}^{D}|x_{i}| + \prod_{i = 1}^{n}|x_{i}| \nonumber
\end{eqnarray}
subject to  $-100 \leq x_{i} \leq 100$.  The global minima is located at $\textbf{x}^{*} = f(0,\cdots,0)$,
$f(\textbf{x}^{*}) = 0$.  \\

\item \textbf{Schwefel 2.23 Function} \cite{SCHWEFEL1981} (Continuous, Differentiable, Non-Separable, Scalable, Unimodal)
\begin{eqnarray}
	       f_{125}(\textbf{x}) &=&  \sum_{i=1}^{D}x_{i}^{10} \nonumber
\end{eqnarray}
subject to  $-10 \leq x_{i} \leq 10$.  The global minima is located at $\textbf{x}^{*} = f(0,\cdots,0)$,
$f(\textbf{x}^{*}) = 0$.  \\

\item \textbf{Schwefel 2.23 Function} \cite{SCHWEFEL1981} (Continuous, Differentiable, Non-Separable, Scalable, Unimodal)
\begin{eqnarray}
	       f_{126}(\textbf{x}) &=&  \sum_{i=1}^{D}x_{i}^{10} \nonumber
\end{eqnarray}
subject to  $-10 \leq x_{i} \leq 10$.  The global minima is located at $\textbf{x}^{*} = f(0,\cdots,0)$,
$f(\textbf{x}^{*}) = 0$.  \\

\item \textbf{Schwefel 2.25 Function} \cite{SCHWEFEL1981} (Continuous, Differentiable, Separable, Non-Scalable, Multimodal)
\begin{eqnarray}
    f_{127}(\textbf{x}) &=&  \sum_{i=2}^{D}(x_{i} - 1)^2 + (x_1 - x_i^2)^2 \nonumber
\end{eqnarray}
subject to $0 \leq x_{i} \leq 10$.  The global minima is located at $\textbf{x}^{*} = f(1,\cdots,1)$,
$f(\textbf{x}^{*}) = 0$.\\

\item \textbf{Schwefel 2.26 Function} \cite{SCHWEFEL1981} (Continuous, Differentiable, Separable, Scalable, Multimodal)
\begin{eqnarray}
    f_{128}(\textbf{x}) &=& -\frac{1}{D}\sum_{i=1}^{D}x_{i}\sin{\sqrt{|x_{i}|}} \nonumber
\end{eqnarray}
subject to $-500 \leq x_{i} \leq 500$.  The global minimum is located at $\textbf{x}^{*} = \pm[\pi(0.5+k)]^2$,
$f(\textbf{x}^{*}) = -418.983$.\\

\item \textbf{Schwefel 2.36 Function} \cite{SCHWEFEL1981} (Continuous, Differentiable, Separable, Scalable, Multimodal)
\begin{eqnarray}
    f_{129}(\textbf{x}) &=& -x_1x_2(72 - 2x_1 -2x_2) \nonumber
\end{eqnarray}
subject to $0 \leq x_{i} \leq 500$.  The global minimum is located at $\textbf{x}^{*} = f(12,\cdots,12)$,
$f(\textbf{x}^{*}) = -3456$.\\

\item \textbf{Shekel 5} \cite{OPACIC1973} (Continuous, Differentiable, Non-Separable, Scalable, Multimodal)
\begin{eqnarray}
    f_{130}(\textbf{x}) &=& - \sum\limits_{i = 1}^5 {\frac{1}{{\sum\limits_{j = 1}^4 {{{\left( {{x_j} - {a_{ij}}} \right)}^2} + {c_i}} }}} \nonumber
\end{eqnarray}
where 
$\textbf{A} = [A_{ij}] = \left[ {\begin{matrix}
4&4&4&4\\
1&1&1&1\\
8&8&8&8\\
6&6&6&6\\
3&7&3&7
\end{matrix}} \right]$, $\textbf{\underline{c}} = {c_i} = \left[ {\begin{matrix}
{0.1}\\
{0.2}\\
{0.2}\\
{0.4}\\
{0.4}
\end{matrix}} \right]$ \\

subject to $0 \leq x_{j} \leq 10$.  The global minima is located at $\textbf{x}^{*} = f(4, 4, 4, 4)$, $f(\textbf{x}^{*}) \approx -10.1499$.  \\

\item \textbf{Shekel 7} \cite{OPACIC1973} (Continuous, Differentiable, Non-Separable, Scalable, Multimodal)
\begin{eqnarray}
    f_{131}(\textbf{x}) &=& - \sum\limits_{i = 1}^7 {\frac{1}{{\sum\limits_{j = 1}^4 {{{\left( {{x_j} - {a_{ij}}} \right)}^2} + {c_i}} }}} \nonumber
\end{eqnarray}
where 
$\textbf{A} = [A_{ij}] = \left[ {\begin{matrix}
4&4&4&4\\
1&1&1&1\\
8&8&8&8\\
6&6&6&6\\
3&7&3&7\\
2&9&2&9\\
5&5&3&3
\end{matrix}} \right]$, $\textbf{\underline{c}} = {c_i} = \left[ {\begin{matrix}
{0.1}\\
{0.2}\\
{0.2}\\
{0.4}\\
{0.4}\\
{0.6}\\
{0.3}
\end{matrix}} \right]$ \\

subject to $0 \leq x_{j} \leq 10$.  The global minima is located at $\textbf{x}^{*} = f(4, 4, 4, 4)$, $f(\textbf{x}^{*}) \approx -10.3999$.  \\

\item \textbf{Shekel 10} \cite{OPACIC1973} (Continuous, Differentiable, Non-Separable, Scalable, Multimodal)
\begin{eqnarray}
    f_{132}(\textbf{x}) &=& - \sum\limits_{i = 1}^{10} {\frac{1}{{\sum\limits_{j = 1}^4 {{{\left( {{x_j} - {a_{ij}}} \right)}^2} + {c_i}} }}} \nonumber
\end{eqnarray}
where 
$\textbf{A} = [A_{ij}] = \left[ {\begin{matrix}
4&4&4&4\\
1&1&1&1\\
8&8&8&8\\
6&6&6&6\\
3&7&3&7\\
2&9&2&9\\
5&5&3&3 \\
8&1&8&1\\
6&2&6&2 \\
7&3.6&7&3.6
\end{matrix}} \right]$, $\textbf{\underline{c}} = {c_i} = \left[ {\begin{matrix}
{0.1}\\
{0.2}\\
{0.2}\\
{0.4}\\
{0.4}\\
{0.6}\\
{0.3}\\
{0.7}\\
{0.5}\\
{0.5}
\end{matrix}} \right]$ \\

subject to $0 \leq x_{j} \leq 10$.  The global minima is located at $\textbf{x}^{*} = f(4, 4, 4, 4)$, $f(\textbf{x}^{*}) \approx -10.5319$.  \\

\item \textbf{Shubert Function} \cite{HENNART1982} (Continuous, Differentiable, Separable?, Non-Scalable, Multimodal)
\begin{eqnarray}
	       f_{133}(\textbf{x}) &=&  \prod_{i=1}^{n}\left(\sum_{j=1}^{5}\textrm{cos}((j + 1)x_{i} + j)\right) \nonumber
\end{eqnarray}
    subject to  $-10 \leq x_{i} \leq 10$, $i \in {1,2,\cdots,n}$.  The 18 global minima are located at
    $\textbf{x}^{*} = f(\{-7.0835, ~~4.8580\},    ~~\{-7.0835, -7.7083\},\\
          ~~~~~~~~~~~~~~~~~~~~~~~~\{-1.4251, -7.0835\},     ~~ \{~~5.4828,  ~~4.8580\},  \\
          ~~~~~~~~~~~~~~~~~~~~~~~~\{-1.4251, -0.8003\},     ~~ \{~~4.8580,  ~~5.4828\},  \\
          ~~~~~~~~~~~~~~~~~~~~~~~~\{-7.7083, -7.0835\},     ~~ \{-7.0835, -1.4251\}, \\
          ~~~~~~~~~~~~~~~~~~~~~~~~\{-7.7083, -0.8003\},     ~~ \{-7.7083,  ~~5.4828\}, \\
          ~~~~~~~~~~~~~~~~~~~~~~~~\{-0.8003, -7.7083\},     ~~ \{-0.8003, -1.4251\},  \\
          ~~~~~~~~~~~~~~~~~~~~~~~~\{-0.8003,~~4.8580\},     ~~ \{-1.4251,  ~~5.4828\}, \\
          ~~~~~~~~~~~~~~~~~~~~~~~~\{~~5.4828, -7.7083\},    ~~ \{~~4.8580, -7.0835\},  \\
          ~~~~~~~~~~~~~~~~~~~~~~~~\{~~5.4828, -1.4251\},    ~~ \{~~4.8580,  -0.8003\})$, \\
          $f(\textbf{x}^{*}) \simeq -186.7309 $.\\

\item \textbf{Shubert 3 Function} \cite{ADORIO2005} (Continuous, Differentiable, Separable, Non-Scalable, Multimodal)
\begin{eqnarray}
	       f_{134}(\textbf{x}) &=&  \left(\sum_{i=1}^{D}\sum_{j=1}^{5}j{\sin}((j + 1)x_{i} + j)\right) \nonumber
\end{eqnarray}
    subject to  $-10 \leq x_{i} \leq 10$.  The global minimum is $f(\textbf{x}^{*}) \simeq -29.6733337 $ with multiple solutions.\\

\item \textbf{Shubert 4 Function} \cite{ADORIO2005} (Continuous, Differentiable, Separable, Non-Scalable, Multimodal)
\begin{eqnarray}
	       f_{135}(\textbf{x}) &=&  \left(\sum_{i=1}^{D}\sum_{j=1}^{5}j{\cos}((j + 1)x_{i} + j)\right) \nonumber
\end{eqnarray}
    subject to  $-10 \leq x_{i} \leq 10$.  The global minimum is $f(\textbf{x}^{*}) \simeq -25.740858 $ with multiple solutions.\\

\item \textbf{Schaffer F6 Function} \cite{SCHAFFER1989} (Continuous, Differentiable, Non-Separable, Scalable, Multimodal)
\begin{eqnarray}
	       f_{136}(\textbf{x}) &=& \sum_{i=1}^{D} {0.5} + \frac{\sin^2{\sqrt{x_{i}^2 + x_{i+1}^2}} - 0.5}{\Big[1 + 0.001(x_{i}^2 + x_{i+1}^2)\Big]^2} \nonumber
        \end{eqnarray}
    subject to  $-100 \leq x_{i} \leq 100$.  The global minimum is located at $\textbf{x}^{*} = f(0,\cdots,0)$, $f(\textbf{x}^{*}) = 0 $.  \\

\item \textbf{Sphere Function} \cite{SCHUMER1968} (Continuous, Differentiable, Separable, Scalable, Multimodal)
\begin{eqnarray}
	f_{137}(\textbf{x}) &=& \sum_{i=1}^{D}x_{i}^{2} \nonumber
\end{eqnarray}
subject to  $0 \leq x_{i} \leq 10$. The global minima is located $\textbf{x}^{*} = f(0,\cdots,0)$,
$f(\textbf{x}^{*}) = 0$. \nonumber \\

\item \textbf{Step Function} (Discontinuous, Non-Differentiable, Separable, Scalable, Unimodal)
\begin{eqnarray}
	f_{138}(\textbf{x}) &=& \sum_{i=1}^{D}\left(\lfloor{|x_{i}|}\rfloor\right) \nonumber
\end{eqnarray}
subject to  $-100 \leq x_{i} \leq 100$. The global minima is located $\textbf{x}^{*} = f(0,\cdots,0) = 0$,
$f(\textbf{x}^{*}) = 0$. \nonumber \\

\item \textbf{Step 2 Function} \cite{BAECK1993} (Discontinuous, Non-Differentiable, Separable, Scalable, Unimodal)
\begin{eqnarray}
	f_{139}(\textbf{x}) &=& \sum_{i=1}^{D}\left(\lfloor{x_{i} + 0.5}\rfloor\right)^2 \nonumber
\end{eqnarray}
subject to  $-100 \leq x_{i} \leq 100$. The global minima is located $\textbf{x}^{*} = f(0.5,\cdots,0.5) = 0$,
$f(\textbf{x}^{*}) = 0$. \nonumber \\

\item \textbf{Step 3 Function} (Discontinuous, Non-Differentiable, Separable, Scalable, Unimodal)
\begin{eqnarray}
	f_{140}(\textbf{x}) &=& \sum_{i=1}^{D}\left(\lfloor{x_{i}^2}\rfloor\right) \nonumber
\end{eqnarray}
subject to  $-100 \leq x_{i} \leq 100$. The global minima is located $\textbf{x}^{*} = f(0,\cdots,0) = 0$,
$f(\textbf{x}^{*}) = 0$. \nonumber \\

\item \textbf{Stepint Function} (Discontinuous, Non-Differentiable, Separable, Scalable, Unimodal)
\begin{eqnarray}
	f_{141}(\textbf{x}) &=& 25 + \sum_{i=1}^{D}\left(\lfloor{x_{i}}\rfloor\right) \nonumber
\end{eqnarray}
subject to  $-5.12 \leq x_{i} \leq 5.12$. The global minima is located $\textbf{x}^{*} = f(0,\cdots,0)$,
$f(\textbf{x}^{*}) = 0$. \nonumber \\

\item \textbf{Streched V Sine Wave Function} \cite{SCHAFFER1989} (Continuous, Differentiable, Non-Separable, Scalable, Unimodal)
\begin{eqnarray}
	f_{142}(\textbf{x}) &=& \sum_{i=1}^{D-1}(x_{i+1}^{2} + x_{i}^{2})^{0.25}\Big[\sin^2\{50(x_{i+1}^{2} + x_{i}^{2})^{0.1}\} + 0.1\Big] \nonumber
\end{eqnarray}
subject to  $-10 \leq x_{i} \leq 10$. The global minimum is located $\textbf{x}^{*} = f(0, 0)$,
$f(\textbf{x}^{*}) = 0$. \nonumber \\

\item \textbf{Sum Squares Function} \cite{HEDAR} (Continuous, Differentiable, Separable, Scalable, Unimodal)
\begin{eqnarray}
	f_{143}(\textbf{x}) &=& \sum_{i=1}^{D}ix_{i}^{2} \nonumber
\end{eqnarray}
subject to  $-10 \leq x_{i} \leq 10$. The global minima is located $\textbf{x}^{*} = f(0,\cdots,0)$,
$f(\textbf{x}^{*}) = 0$. \nonumber \\

\item \textbf{Styblinski-Tang Function} \cite{SILAGADZE2007} (Continuous, Differentiable, Non-Separable, Non-Scalable, Multimodal)
\begin{eqnarray}
	       f_{144}(\textbf{x}) &=&  \frac{1}{2}\sum_{i=1}^{n}(x_{i}^{4} - 16x_{i}^{2} + 5x_{i}) \nonumber
\end{eqnarray}
    subject to  $-5 \leq x_{i} \leq 5$. The global minimum is located $\textbf{x}^{*} = f(-2.903534,-2.903534)$, $f(\textbf{x}^{*}) = -78.332$.

\item \textbf{Table 1 / Holder Table 1 Function} \cite{MISHRA2006_6} (Continuous, Differentiable, Separable, Non-Scalable, Multimodal)
    \begin{eqnarray}
        f_{145}(\textbf{x}) &=& -|\textrm{cos}(x_{1})\textrm{cos}(x_{2})e^{|1 - (x_{1} + x_{2})^{0.5}/\pi|}|    \nonumber
    \end{eqnarray}
    subject to  $-10 \leq x_{i} \leq 10$.

    The four global minima are located at $\textbf{x}^{*}$ = $f(\pm 9.646168$, $\pm 9.646168)$,
    $f(\textbf{x}^{*}) = -26.920336 $.  \\

\item \textbf{Table 2 / Holder Table 2 Function} \cite{MISHRA2006_6} (Continuous, Differentiable, Separable, Non-Scalable, Multimodal)
    \begin{eqnarray}
        f_{146}(\textbf{x}) &=& -|\textrm{sin}(x_{1})\textrm{cos}(x_{2})e^{|1 - (x_{1} + x_{2})^{0.5}/\pi|}|    \nonumber
    \end{eqnarray}
    subject to  $-10 \leq x_{i} \leq 10$.

    The four global minima are located at
    $\textbf{x}^{*}$ = $f(\pm 8.055023472141116$, $\pm 9.664590028909654)$,
    $f(\textbf{x}^{*})$ = $-19.20850 $.  \\

\item \textbf{Table 3 / Carrom Table Function} \cite{MISHRA2006_6} (Continuous, Differentiable, Non-Separable, Non-Scalable, Multimodal)
        \begin{eqnarray}
	       f_{147}(\textbf{x}) &=& -[(\textrm{cos}(x_{1})\textrm{cos}(x_{2}) \nonumber \\
                                           & & \exp|1 - [(x_{1}^{2} + x_{2}^{2})^{0.5}]/\pi|)^{2}]/30 \nonumber
        \end{eqnarray}
    subject to  $-10 \leq x_{i} \leq 10$.

    The four global minima are located at $\textbf{x}^{*}$ = $f(\pm 9.646157266348881$, $\pm 9.646134286497169)$,
    $f(\textbf{x}^{*}) = -24.1568155$.  \\

\item \textbf{Testtube Holder Function} \cite{MISHRA2006_6} (Continuous, Differentiable, Separable, Non-Scalable, Multimodal)
\begin{eqnarray}
	       f_{148}(\textbf{x}) &=& -4\big[(\textrm{sin}(x_{1})\textrm{cos}(x_{2}) \nonumber \\
                                           & & e^{|\textrm{cos}[(x_{1}^{2} + x_{2}^{2})/{200}]|})\big] \nonumber
\end{eqnarray}
    subject to  $-10 \leq x_{i} \leq 10$.  The two global minima are located at $\textbf{x}^{*} = f(\pm \pi/2, 0)$, $f(\textbf{x}^{*}) = -10.872300$.  \\

\item \textbf{Trecanni Function} \cite{DIXON1978} (Continuous, Differentiable, Separable, Non-Scalable, Unimodal)
\begin{eqnarray}
	       f_{149}(\textbf{x}) &=&  x_{1}^{4} - 4x_{1}^{3} + 4x_{1} + x_{2}^2 \nonumber
\end{eqnarray}
    subject to  $-5 \leq x_{i} \leq 5$.  The two global minima are located at $\textbf{x}^{*} = f(\{0,0\}, \{-2, 0\})$, $f(\textbf{x}^{*}) = 0$.

\item \textbf{Trid 6 Function} \cite{HEDAR} (Continuous, Differentiable, Non-Separable, Non-Scalable, Multimodal)
\begin{eqnarray}
	       f_{150}(\textbf{x}) &=&  \sum_{i=1}^{D}\left(x_{i} - 1 \right)^{2} - \sum_{i=1}^{D}x_{i}x_{i-1} \nonumber
\end{eqnarray}
    subject to  $-6^2 \leq x_{i} \leq 6^2$.  The global minima is located at $f(\textbf{x}^{*}) = -50$. \nonumber \\
\item \textbf{Trid 10 Function} \cite{HEDAR} (Continuous, Differentiable, Non-Separable, Non-Scalable, Multimodal)
\begin{eqnarray}
	       f_{151}(\textbf{x}) &=&  \sum_{i=1}^{D}\left(x_{i} - 1 \right)^{2} - \sum_{i=1}^{D}x_{i}x_{i-1} \nonumber
\end{eqnarray}
    subject to  $-100 \leq x_{i} \leq 100$.  The global minima is located at $f(\textbf{x}^{*}) = -200$. \nonumber \\

\item \textbf{Trefethen Function} \cite{ADORIO2005} (Continuous, Differentiable, Non-Separable, Non-Scalable, Multimodal)
\begin{eqnarray}
	       f_{152}(\textbf{x}) &=&  e^{\textrm{sin}(50x_{1})} + \textrm{sin}(60e^{x_{2}}) \nonumber \\
                                           & & + \textrm{sin}(70\textrm{sin}(x_{1})) + \textrm{sin}(\textrm{sin}(80x_{2})) \nonumber \\
    		                               & &  - \textrm{sin}(10(x_{1} + x_{2})) + \frac{1}{4}(x_{1}^{2} + x_{2}^{2}) \nonumber
\end{eqnarray}
    subject to  $-10 \leq x_{i} \leq 10$.  The global minimum is located at $\textbf{x}^{*} = f(-0.024403,0.210612)$, $f(\textbf{x}^{*}) = -3.30686865$.

\item \textbf{Trigonometric 1 Function} \cite{DIXON1978} (Continuous, Differentiable, Non-Separable, Scalable, Multimodal)

\begin{eqnarray}
	       f_{153}(\textbf{x}) &=&  \sum_{i=1}^{D}[D - \sum_{j=1}^{D}\cos{x_{j}} \nonumber \\
                                           & & + i(1 - \cos(x_{i}) - \sin(x_{i}))]^{2} \nonumber
\end{eqnarray}
subject to  $0 \leq x_{i} \leq pi$.  The global minimum is located at $\textbf{x}^{*} = f(0,\cdots,0)$, $f(\textbf{x}^{*}) = 0$ \nonumber \\

\item \textbf{Trigonometric 2 Function} \cite{FU2006} (Continuous, Differentiable, Non-Separable, Scalable, Multimodal)
\begin{eqnarray}
	       f_{154}(\textbf{x}) &=&  1 + \sum_{i=1}^{D} 8\sin^2\big[7(x_i - 0.9)^2\big] + 6\sin^2\big[14(x_1 - 0.9)^2\big] + (x_i - 0.9)^2\nonumber
\end{eqnarray}
subject to  $-500 \leq x_{i} \leq 500$.  The global minimum is located at $\textbf{x}^{*} = f(0.9,\cdots,0.9)$, $f(\textbf{x}^{*}) = 1$ \nonumber \\

\item \textbf{Tripod Function} \cite{RAHNAMAYAN2007} (Discontinuous, Non-Differentiable, Non-Separable, Non-Scalable, Multimodal)
\begin{eqnarray}
	       f_{155}(\textbf{x}) &=& p(x_{2})(1 + p(x_{1})) \nonumber \\
                                           & & + |x_{1} + 50p(x_{2})(1-2p(x_{1}))| \nonumber \\
                                           & & + |x_{2} + 50(1 - 2p(x_{2}))| \nonumber
\end{eqnarray}
    subject to  $-100 \leq x_{i} \leq 100$, where $ p(x) = 1 $ for $x \geq 0$.  The global minimum is located at $\textbf{x}^{*} = f(0,-50)$, $f(\textbf{x}^{*}) = 0$. \\

\item \textbf{Ursem 1 Function} \cite{ROENKKOENEN2009} (Separable)
\begin{eqnarray}
	       f_{156}(\textbf{x}) &=& -\textrm{sin}(2x_{1} - 0.5\pi) - 3\textrm{cos}(x_{2}) - 0.5x_{1} \nonumber
\end{eqnarray}
    subject to  $-2.5 \leq x_{1} \leq 3$ and $-2 \leq x_{2} \leq 2$, and has single global and local minima.

\item \textbf{Ursem 3 Function} \cite{ROENKKOENEN2009}(Non-separable)
\begin{eqnarray}
	       f_{157}(\textbf{x}) &=& -\textrm{sin}(2.2\pi x_{1} + 0.5\pi).\frac{2 - \left|x_{1}\right|}{2}.\frac{3 - \left|x_{1}\right|}{2} \nonumber \\
& & -\textrm{sin}(0.5\pi x_{2}^2 + 0.5\pi).\frac{2 - \left|x_{2}\right|}{2}.\frac{3 - \left|x_{2}\right|}{2}\nonumber
\end{eqnarray}
    subject to  $-2 \leq x_{1} \leq 2$ and $-1.5 \leq x_{2} \leq 1.5$, and has single global minimum and four regularly spaced local minima
    positioned in a direct line, such that global minimum is in the middle.

\item \textbf{Ursem 4 Function} \cite{ROENKKOENEN2009} (Non-separable)

\begin{eqnarray}
	       f_{158}(\textbf{x}) &=& -3\textrm{sin}(0.5\pi x_{1} + 0.5\pi).\frac{2 - \sqrt{x_{1}^2 + x_{2}^2}}{4} \nonumber
\end{eqnarray}
    subject to  $-2 \leq x_{i} \leq 2$, and has single global minimum positioned at the middle and four local minima
    at the corners of the search space.

\item \textbf{Ursem Waves Function} \cite{ROENKKOENEN2009}(Non-separable)
\begin{eqnarray}
	       f_{159}(\textbf{x}) &=& -0.9x_{1}^2 + (x_{2}^2 - 4.5x_{2}^2)x_{1}x_{2} \nonumber \\
                                           & & +4.7\textrm{cos}(3x_{1} - x_{2}^2(2 + x_{1}))\textrm{sin}(2.5\pi x_{1}) \nonumber
\end{eqnarray}
    subject to  $-0.9 \leq x_{1} \leq 1.2$ and $-1.2 \leq x_{2} \leq 1.2$, and has single global minimum and nine irregularly spaced
    local minima in the search space.

\item \textbf{Venter Sobiezcczanski-Sobieski Function} \cite{BEGAMBRE2009} (Continuous, Differentiable, Separable, Non-Scalable)
\begin{eqnarray}
	       f_{160}(\textbf{x}) &=& x_{1}^{2} - 100\textrm{cos}(x_1)^{2} \nonumber \\
                                           & & - 100\textrm{cos}(x_{1}^{2}/30) + x_{2}^{2} \nonumber \\
                                           & & - 100\textrm{cos}(x_{2})^{2} - 100\textrm{cos}(x_{2}^{2}/30) \nonumber
\end{eqnarray}
    subject to  $-50 \leq x_{i} \leq 50$.  The global minimum is located at $\textbf{x}^{*} = f(0,0)$, $f(\textbf{x}^{*}) = -400$. \\

\item \textbf{Watson Function} \cite{SCHWEFEL1981} (Continuous, Differentiable, Non-Separable, Scalable, Unimodal)
\begin{eqnarray}
  f_{161}(\textbf{x}) &=& \sum_{i = 0} ^{29}   \{ \sum_{j = 0}^4 ((j-1) a_i^j x_{j+1}) - \left[ \sum_{j = 0}^5 a_i^j x_{j+1} \right]^2 - 1\}^2 + x_1^2 \nonumber
\end{eqnarray}
subject to $|x_i| \le 10$, where the coefficient $a_i = i / 29.0$. The global minimum is located at $\textbf{x}^{*} = f(-0.0158, 1.012, -0.2329, 1.260, -1.513, 0.9928)$,
$f(\textbf{x}^{*}) = 0.002288$.\\

\item \textbf{Wayburn Seader 1 Function} \cite{WAYBURN1987} (Continuous, Differentiable, Non-Separable, Scalable, Unimodal)
\begin{eqnarray}
  f_{162}(\textbf{x}) &=& (x_1^6 + x_2^4 - 17)^2 + (2x_1 + x_2 - 4)^2 \nonumber
\end{eqnarray}
The global minimum is located at $\textbf{x}^{*} = f\{(1,2), (1.597, 0.806)\}$,
$f(\textbf{x}^{*}) = 0$.\\

\item \textbf{Wayburn Seader 2 Function} \cite{WAYBURN1987} (Continuous, Differentiable, Non-Separable, Scalable, Unimodal)
\begin{eqnarray}
  f_{163}(\textbf{x}) &=& \Big[1.613 - 4(x_1 - 0.3125)^2 - 4(x_2 - 1.625)^2\Big]^2 + (x_2 - 1)^2 \nonumber
\end{eqnarray}
subject to $-500 \leq 500$.  The global minimum is located at $\textbf{x}^{*} = f\{(0.2, 1), (0.425, 1)\}$,
$f(\textbf{x}^{*}) = 0$.\\

\item \textbf{Wayburn Seader 3 Function} \cite{WAYBURN1987} (Continuous, Differentiable, Non-Separable, Scalable, Unimodal)
\begin{eqnarray}
  f_{164}(\textbf{x}) &=& 2\frac{x_1^3}{3} - 8x_1^2 + 33x_1 - x_1x_2 + 5 + \Big[(x_1 - 4)^2 + (x_2 - 5)^2 - 4\Big]^2 \nonumber
\end{eqnarray}
subject to $-500 \leq 500$.  The global minimum is located at $\textbf{x}^{*} = f(5.611, 6.187)$,
$f(\textbf{x}^{*}) = 21.35$.\\

\item \textbf{W / Wavy Function} \cite{COURRIEU1997} (Continuous, Differentiable, Separable, Scalable, Multimodal)
\begin{eqnarray}
	f_{165}(\textbf{x}) &=& 1 - \frac{1}{D}\sum\limits_{i = 1}^D \textrm{cos}(kx_{i})e^\frac{-x_{i}^2}{2}\nonumber
\end{eqnarray}
subject to  $-\pi \leq x_{i} \leq \pi$.  The global minimum is located at $\textbf{x}^{*} = f(0,0)$, $ f(\textbf{x}^{*})= 0$.\\
The number of local minima is $kn$ and $(k+1)n$ for odd and even $k$ respectively.  For $D=2$ and $k=10$, there are 121 local minima.

\item \textbf{Weierstrass Function} \cite{SUGANTHAN2005}(Continuous, Differentiable, Separable, Scalable, Multimodal)
\begin{eqnarray}
	       f_{166}(\textbf{x}) &=& \sum_{i=1}^{n}[\sum_{k=0}^{kmax}a^{k}\textrm{cos}(2 \pi b^{k}(x_{i} + 0.5)) \nonumber \\
                                           & & - n\sum_{k=0}^{kmax}a^{k}\textrm{cos}(\pi b^{k})] \nonumber
\end{eqnarray}
    subject to  $-0.5 \leq x_{i} \leq 0.5$.  The global minima is located at $\textbf{x}^{*} = f(0,\cdots,0)$, $f(\textbf{x}^{*}) = 0$. \\


\item \textbf{Whitley Function} \cite{WHITLEY1996} (Continuous, Differentiable, Non-Separable, Scalable, Multimodal)

\begin{eqnarray}
    f_{167}(\textbf{x}) &=& \sum_{i=1}^D\sum_{j=1}^D\big[\frac{(100(x_{i}^2 - x_{j})^2 + (1 - x_{j})^2)^2}{4000} \nonumber \\
    & & - \textrm{cos}\left(100(x_{i}^2 - x_{j})^2 + (1 - x_{j})^2 + 1\right)\big]\nonumber
\end{eqnarray}
combines a very steep overall slope with a highly multimodal area around the global minimum located at $x_{i}=1$, where $i = 1,...,D$.

\item \textbf{Wolfe Function} \cite{SCHWEFEL1981} (Continuous, Differentiable, Separable, Scalable, Multimodal)
\begin{eqnarray}
	       f_{168}(\textbf{x}) &=& \frac{4}{3}(x_1^2 + x_2^2 - x_{1}x_{2})^0.75 + x_3
\end{eqnarray}
    subject to  $0 \leq x_{i} \leq 2$.  The global minima is located at $\textbf{x}^{*} = f(0,\cdots,0)$, $f(\textbf{x}^{*}) = 0$. \\

\item \textbf{Xin-She Yang (Function 1)} (Separable)\\
This is a generic stochastic and non-smooth function proposed in \cite{YANG2010a,Yang2010b}.
\begin{eqnarray}
            f_{169}(\textbf{x}) &=& \sum_{i=1}^{D}\epsilon_{i}\abs{x_{i}}^{i} \nonumber
\end{eqnarray}

subject to  $-5 \leq x_{i} \leq 5$.  The variable $\epsilon_{i}, (i=1,2,\cdots,D)$ is a random variable uniformly distributed in $[0,1]$.
The global minima is located at $\textbf{x}^{*} = f(0,\cdots,0)$, $f(\textbf{x}^{*}) = 0$.  \\

\item \textbf{Xin-She Yang (Function 2)}(Non-separable)
\begin{eqnarray}
            f_{170}(\textbf{x}) &=& \big(\sum_{i=1}^{D}\abs{x_{i}}\big)\exp\big[-\sum_{i=1}^{D}\textrm{sin}(x_{i}^2)\big] \nonumber
\end{eqnarray}

subject to  $-2\pi \leq x_{i} \leq 2\pi$.  The global minima is located at $\textbf{x}^{*} = f(0,\cdots,0)$, $f(\textbf{x}^{*}) = 0$.  \\

\item \textbf{Xin-She Yang (Function 3)} (Non-separable)
\begin{eqnarray}
	       f_{171}(\textbf{x}) &=& \left[e^{-\sum_{i=1}^{D}(x_{i}/\beta)^{2m}} - 2e^{-\sum_{i=1}^{D}(x_{i})^{2}}.\prod_{i=1}^{D}\cos^2(x_{i})\right]\nonumber
\end{eqnarray}
subject to  $-20 \leq x_{i} \leq 20$.  The global minima for $m=5$ and $\beta=15$ is located at $\textbf{x}^{*} = f(0,\cdots,0)$, $f(\textbf{x}^{*}) = -1$. \\

\item \textbf{Xin-She Yang (Function 4)} (Non-separable)
\begin{eqnarray}
	       f_{172}(\textbf{x}) &=& \left[{\sum_{i=1}^{D} \sin^{2}(x_{i})} - e^{-\sum_{i=1}^{D}x_{i}^2}\right].e^{-\sum_{i=1}^{D}\sin^2\sqrt{{|x_{i}|}}} \nonumber
\end{eqnarray}
subject to  $-10 \leq x_{i} \leq 10$.  The global minima is located at $\textbf{x}^{*} = f(0,\cdots,0)$, $f(\textbf{x}^{*}) = -1$. \\

\item \textbf{Zakharov Function} \cite{RAHNAMAYAN2007} (Continuous, Differentiable, Non-Separable, Scalable, Multimodal)
\begin{eqnarray}
	       f_{173}(\textbf{x}) &=& \sum_{i=1}^{n} x_{i}^{2} + \left(\frac{1}{2}\sum_{i=1}^{n}ix_{i}\right)^2
                                            + \left(\frac{1}{2}\sum_{i=1}^{n}ix_{i}\right)^4 \nonumber
\end{eqnarray}
    subject to  $-5 \leq x_{i} \leq 10$.  The global minima is located at $\textbf{x}^{*} = f(0,\cdots,0)$,
    $f(\textbf{x}^{*}) = 0$. \nonumber \\

\item \textbf{Zettl Function} \cite{SCHWEFEL1995} (Continuous, Differentiable, Non-Separable, Non-Scalable, Unimodal)
\begin{eqnarray}
	       f_{174}(\textbf{x}) &=& (x_{1}^{2} + x_{2}^{2} - 2x_{1})^{2} + 0.25 x_{1} \nonumber
\end{eqnarray}
    subject to  $-5 \leq x_{i} \leq 10$.  The global minima is located at $\textbf{x}^{*} = f(-0.0299, 0)$, $f(\textbf{x}^{*}) = -0.003791$. \\

\item \textbf{Zirilli or Aluffi-Pentini's Function} \cite{ALI2005} (Continuous, Differentiable, Separable, Non-Scalable, Unimodal)
\begin{eqnarray}
    f_{175}(\textbf{x}) &=&  0.25x_{1}^{4} - 0.5x_{1}^{2} + 0.1x_{1} + 0.5x_{2}^{2} \nonumber
\end{eqnarray}
subject to  $-10 \leq x_{i} \leq 10$.  The global minimum is located at $\textbf{x}^{*}=(-1.0465,0)$, $f(\textbf{x}^{*}) \approx -0.3523 $. \\

\end{enumerate}

\section{Conclusions}

Test functions are important to validate and compare optimization algorithms, especially newly developed algorithms.
Here, we have attempted to provide the most comprehensive list of known benchmarks or test functions. However,
it is may be possibly that we have missed some functions, but this is not intentional. This list is based on
all the literature known to us by the time of writing.
It can be expected that all these functions should be used for testing new optimization algorithms so as to
provide a more complete view about the performance of any algorithms of interest.